\pdfoutput=1
\documentclass{technical_report}

\usepackage{natbib}

\usepackage[utf8]{inputenc}
\usepackage[T1]{fontenc}
\usepackage{hyperref}
\usepackage{url}
\usepackage{booktabs}
\usepackage{amsfonts}
\usepackage{amsmath}
\usepackage{amssymb}
\usepackage{nicefrac}
\usepackage{microtype}
\usepackage{xcolor}
\usepackage{graphicx}
\usepackage{algorithm}
\usepackage{algpseudocode}
\usepackage{multirow}
\usepackage{subcaption}
\usepackage{enumitem}
\usepackage{colortbl}
\usepackage{makecell}
\usepackage[most]{tcolorbox}
\usepackage{textcomp}
\usepackage{upquote}
\usepackage{pifont}
\usepackage{placeins}
\usepackage{wrapfig}
\usepackage{fancyvrb}
\usepackage{fvextra}

\usepackage[utf8]{inputenc}
\usepackage[T1]{fontenc}
\DeclareUnicodeCharacter{2501}{\textemdash}

\definecolor{deepgreen}{HTML}{007F00}
\newcommand{\cmark}{\textcolor{deepgreen}{\ding{51}}}
\newcommand{\xmark}{\textcolor{red!70!black}{\ding{55}}}

\definecolor{background}{HTML}{f5f5f5}
\definecolor{frame}{HTML}{5581b0}

\DeclareRobustCommand{\umassaffiliation}{%
  \hspace{0.15em}\raisebox{-0.25em}{\includegraphics[height=1.40em]{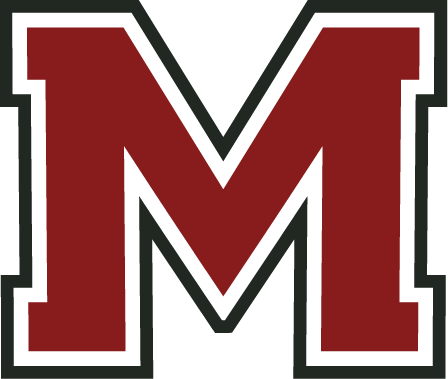}}%
  \hspace{0.28em}\textsf{UMass Amherst}%
}
\DeclareRobustCommand{\zoomaffiliation}{%
  \hspace{0.15em}\raisebox{-0.18em}{\includegraphics[height=1.05em]{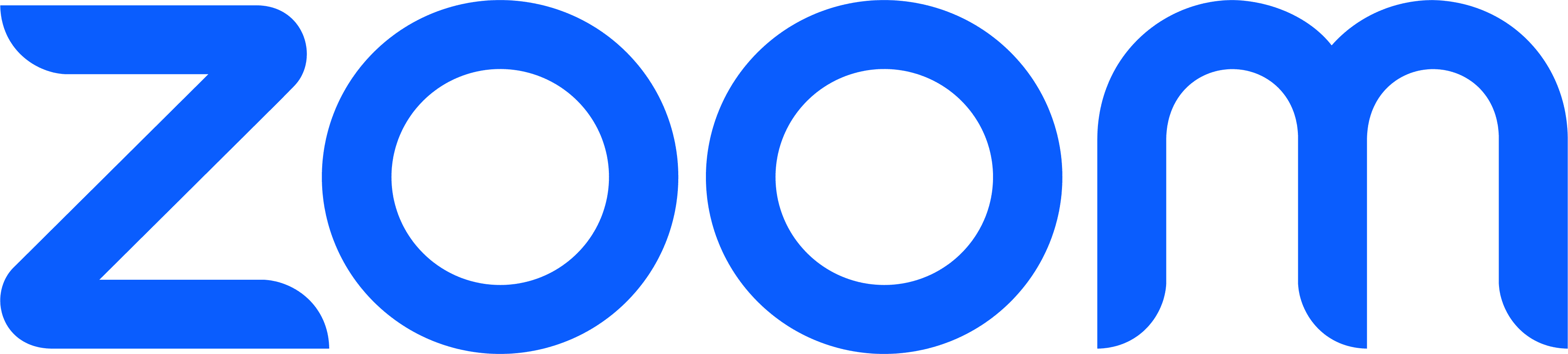}}%
}
\DeclareRobustCommand{\emoryaffiliation}{%
  \hspace{0.15em}\raisebox{-0.30em}{\includegraphics[height=1.50em]{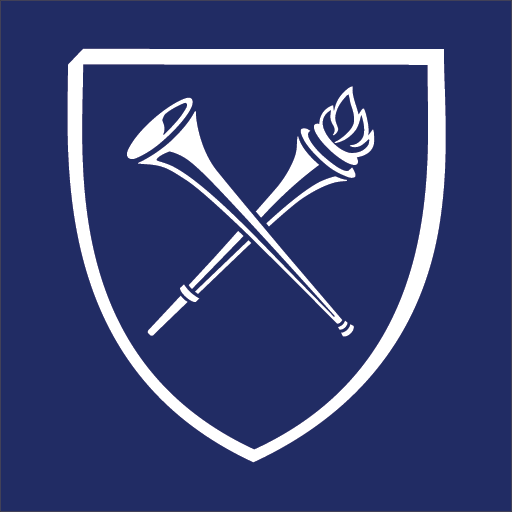}}%
  \hspace{0.28em}\textsf{Emory University}%
}

\DeclareRobustCommand{\hawaiiaffiliation}{%
  \hspace{0.1em}\raisebox{-0.3em}{\includegraphics[height=1.7em]{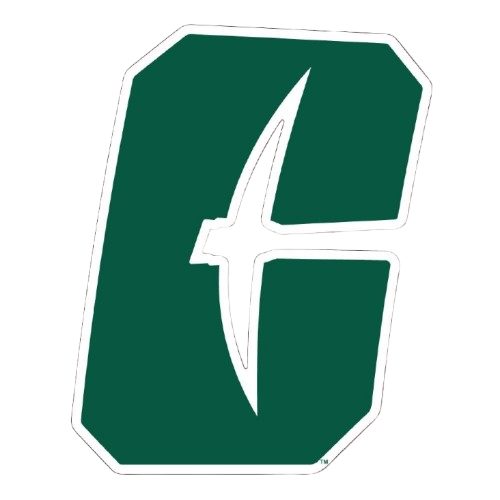}}%
  \hspace{0.15em}\textsf{UNC Charlotte}%
}

\title{An Empirical Study of Harness Design \\for Coding Agents}

\author[1 * \dagger]{Run-Ze Fan}
\author[3 * \dagger]{Zihao Zhang}
\author[2]{Simin Ma}
\author[2]{Yebowen Hu}
\author[4 \dagger]{Shouju Wang}
\author[2]{\protect\\Kaiqiang Song}
\author[3]{Fei Liu}
\author[1]{Hamed Zamani}
\author[2]{Xiaoyang Wang}

\affiliation[1]{\umassaffiliation}
\affiliation[2]{\zoomaffiliation}
\affiliation[3]{\emoryaffiliation}
\affiliation[4]{\hawaiiaffiliation}

\contribution[*]{Equal contribution}
\contribution[\dagger]{Work completed during internships at Zoom Video Communications}
\metadata[Emails]{\email{runzefan@cs.umass.edu}, \email{zihao.zhang@emory.edu}, \email{Xiaoyang.W@zoom.us}}

\begin{document}

\maketitle

\begin{abstract}
Coding harnesses shape how autonomous coding agents translate model capabilities into long-horizon software-engineering performance, yet existing work typically evaluates harnesses as monolithic systems, leaving the effectiveness of individual components unclear. To enable component-level comparisons, we study this question with a lightweight coding harness whose execution loop is fixed while three components are varied: planning, action space, and context management. Across four models evaluated on SWE-Bench Verified and Terminal-Bench 2.1, we evaluate 176 matched settings spanning five context-management strategies, four context-window budgets, and targeted ablations of planning and action space. We find that: (1) Context management becomes increasingly valuable as the context-window budget tightens, with most of its benefit coming from preventing context-overflow failures. (2) Staging rule-based elision before LLM-based summarization provides the strongest overall efficiency among the context-management strategies, whereas making elided content recoverable adds machinery that models rarely use and yields no accuracy gain. (3) Planning shifts from an accuracy scaffold for weaker models to a cost saver for stronger models, with little change in accuracy. (4) Predefined tools improve performance for models with weaker bash proficiency, whereas bash-capable models can operate effectively with a bash-only interface and achieve substantially lower cost, especially on command-line-centric tasks.
Trajectory-level analysis explains these effects: context management extends execution trajectories without substantially altering agent behavior, planning changes where trajectories stop, and the action space changes the granularity at which code is written. These findings inform model- and budget-aware harness design and provide a modular framework for evaluating future harness components.

\end{abstract}


\begin{center}
\includegraphics[height=0.29\textheight,width=\textwidth]{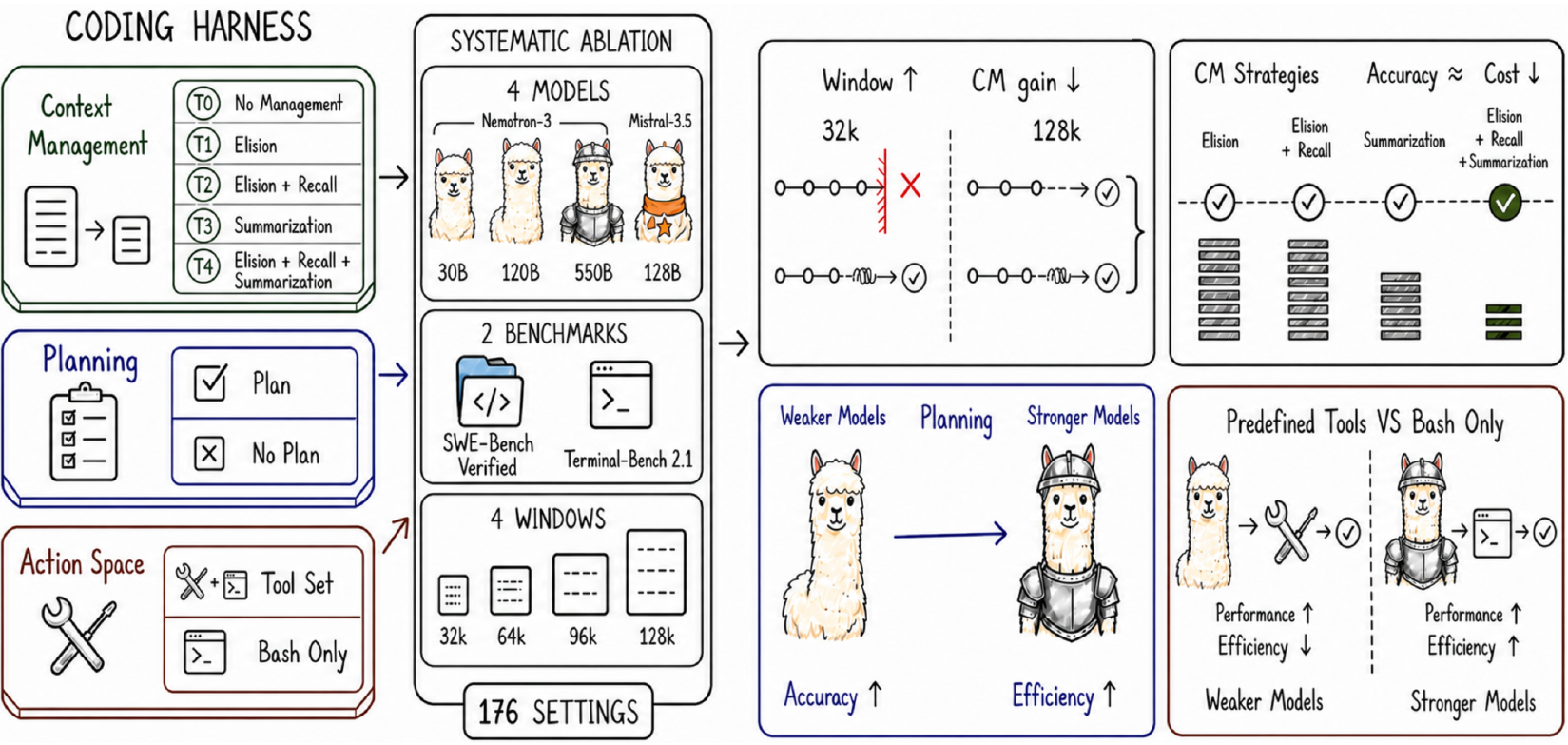}
\captionof{figure}{\textbf{Dissecting the coding harness.} We systematically ablate context management, planning, and the action space, revealing four conditional effects across context budgets, model capabilities, and task types.}
\label{fig:graphicalabstract}
\end{center}


\section{Introduction}


Large language models (LLMs) are increasingly used to resolve real software-engineering tasks autonomously, including closing GitHub issues~\citep{jimenez2024swebench} and completing end-to-end terminal tasks~\citep{merrill2026terminal}. This performance is achieved by having LLMs operate inside a coding harness, a software layer whose components intervene on different aspects of agent behavior: a planning scaffold maintains task structure, an action interface determines how model intentions become executable operations, and a context-management policy decides what interaction history remains available under a finite window~\citep{yang2024sweagent,wang2025openhands,rombaut2026scaffold}. These choices are not incidental implementation details: changing the harness while holding the model fixed can substantially change model performance~\citep{yang2024sweagent,wang2024executable,lewis2026same}.

Despite the empirical success of coding harnesses, many existing studies evaluate them as complete systems~\citep{wang2025openhands,wong2025confucius,xia2024agentless,arora2024masai}. For example, a cross-harness evaluation by \citet{cao2026qwen3coder} reports that Claude-Opus-4.5 performs best with OpenHands among the evaluated harnesses, whereas Claude-Sonnet-4.5 performs best with SWE-Agent, suggesting that harness preferences can vary across models. However, comparisons between complete harnesses conflate multiple mechanisms, so a performance difference between two agents does not reveal whether the gain comes from planning, tool design, context management, or their interaction with the underlying model. This raises a research question: \textbf{Are harness components generally useful across settings, or does each component's effectiveness depend on model capability, task type, and resource budget?} Prior work has examined component interactions and architectural choices across models~\citep{liu2026more,bogavelli2025agentarch,mehtiyev2026beyond,rombaut2026scaffold}, but has not jointly characterized implementation-level planning, workspace action interfaces, and context-management policies on long-horizon coding tasks across both an explicit context-window sweep and a within-family model-scale axis. As a result, existing evidence does not explain which components account for differences across harnesses or when those components transfer.

To address this challenge, we build a coding harness whose surrounding execution loop remains fixed while varying three central components: planning, action space, and context management. We focus on these components because prior systems identify them as complementary requirements of long-horizon coding agents, with planning maintaining task progress~\citep{bairi2024codeplan}, the action space translating model intentions into executable workspace operations~\citep{yang2024sweagent,wang2024executable}, and context management preserving useful information as trajectories grow~\citep{packer2023memgpt,wu2025resum}. Other operational mechanisms, such as permission handling, post-edit diagnostics, and stuck detection, are held fixed to provide a common execution substrate. The \textbf{planning} component maintains an explicit task plan that the model can update throughout a trajectory. The \textbf{action space} exposes either a predefined workspace tool set or a \texttt{bash}-only interface. For \textbf{context management}, we define five strategies. T0 applies no additional cross-turn compaction and terminates when the context window is exceeded. T1 elides stale tool observations. T2 adds external storage and \texttt{recall\_event}, making the elided observations recoverable. T3 uses LLM summarization without elision. T4 combines elision, recoverable external storage, and summarization in a staged policy that applies elision before invoking summarization. This modular design allows us to hold the model, task, execution loop, and unablated components fixed while estimating the conditional effect of each implemented intervention.


Using this harness, we evaluate three sizes of Nemotron-3~\citep{blakeman2025nvidia}, including 30B, 120B, and 550B, as a within-family capability axis, with Mistral-Medium-3.5-128B~\citep{mistral2026medium} as a cross-family comparison. 
We use these models as probes of capability and interaction style, not as permanent optimization targets. 
Our findings therefore provide transferable diagnostics for future models facing the same context, action space, and planning tradeoffs.
We evaluate every model on two complementary long-horizon coding benchmarks: SWE-Bench Verified~\citep{jimenez2024swebench}, which tests repository-level issue resolution, and Terminal-Bench 2.1~\citep{merrill2026terminal}, which tests end-to-end terminal task completion. For context management, we compare five strategies under four context-window budgets of 32k, 64k, 96k, and 128k tokens. We separately ablate planning and the action space at a 128k context-window budget with T4 context management strategy, yielding 176 experimental settings. Beyond success rate and cost, we conduct trajectory-level analysis to characterize how each intervention changes task progression, termination behavior, context use, and tool invocation. Our main findings are summarized below:

\begin{itemize}
    \item \textbf{Context management matters most when the context-window budget is tight.} It prevents context overflow from prematurely terminating execution, allowing agents to progress to code modification and verification. Its accuracy benefit diminishes as the context window expands.
    \item \textbf{Staging elision before LLM summarization (T4) provides the strongest efficiency among the context-management strategies.} T4 maintains mean success similar to the other managed strategies while controlling peak context and reducing reliance on summarization calls. 
    In contrast, the recall mechanism that makes elision reversible is rarely invoked and does not improve accuracy over elision alone.
    \item \textbf{Planning changes from an accuracy scaffold to an efficiency aid as model capability increases.} For weaker models, planning keeps the trajectory alive long enough to attempt an edit, raising success at additional cost; for the stronger models, it mainly removes redundant post-edit verification, lowering cost with only small changes in accuracy.

    \item \textbf{Predefined tools improve performance for bash-weak models, while bash-only interfaces reduce cost for bash-capable models.} Predefined tools reduce reliance on shell commands, whereas bash-capable models can combine multiple operations per call. The resulting accuracy--cost trade-off varies by task type.
\end{itemize}

\section{Harness Design}
\label{sec:harness}

\begin{figure}[t]
\centering
\includegraphics[width=\textwidth]{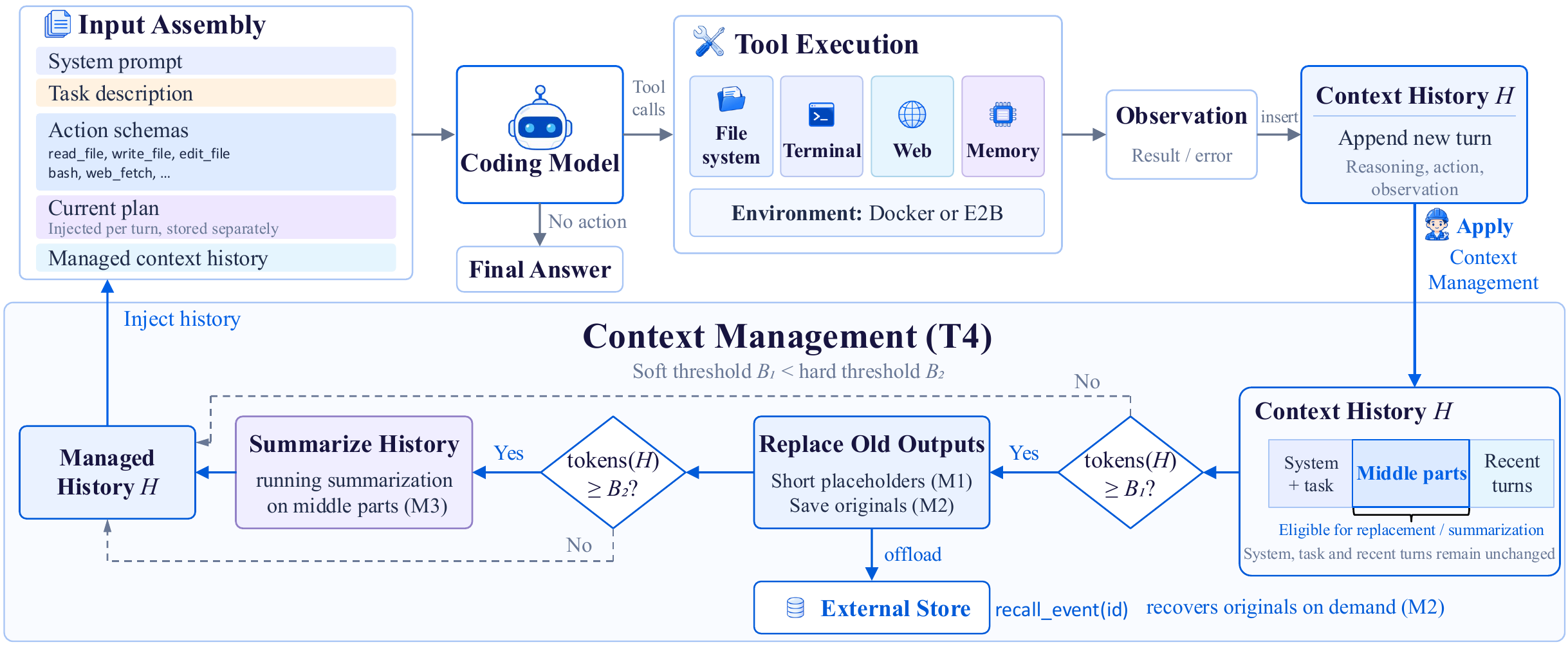}
\caption{\textbf{Overview of the coding harness.} \textbf{Top:} the ReAct loop, in which each turn assembles the model input, executes the emitted tool calls in the task container, and appends the observation to the history $H$. Planning enters through the injected plan, the action space through the exposed schemas, and context management through the history the model sees. \textbf{Bottom:} the T4 strategy. M1--M3 are the three context-management mechanisms. Above the soft threshold $B_1$, bulky middle-region tool outputs are replaced by stubs (M1) and offloaded to an external store recoverable via \texttt{recall\_event} (M2); above the hard threshold $B_2$, the oldest middle events are summarized (M3). The preamble and recent turns stay verbatim.}

\label{fig:harness_overview}
\end{figure}

To examine how the contribution of each harness component varies across models and computational budgets, we build a lightweight harness from scratch.
Many existing harnesses couple implementation choices that are difficult to vary independently. Our modular design allows components to be independently configured and composed, enabling controlled component-level analysis.
The harness follows a ReAct loop~\citep{yao2022react}, with each turn comprising a reasoning step, an action, and an observation (Figure~\ref{fig:harness_overview}).
We vary three components: planning (\S\ref{sec:planning}), the action space (\S\ref{sec:action}), and context management (\S\ref{sec:context}), described below. Other supporting components like, workspace access controls, post-edit diagnostics, and stuck detection remain fixed across ablations to isolate each intervention's effect.

\subsection{Planning}
\label{sec:planning}

Planning provides an explicit, persistent representation of task progress maintained by the model. When enabled, a system instruction defines the protocol (Figure~\ref{prompt:planning_system}), and a first-turn reminder requests an initial plan before action (Figure~\ref{prompt:planning_reminder_first}). The model maintains this plan through the \texttt{update\_plan} tool (Figure~\ref{prompt:tool_update_plan}). Subsequent turns append the plan to the model input without storing it in conversation history (Figure~\ref{prompt:planning_reminder_subsequent}); Appendix~\ref{app:prompts_planning} describes how these blocks are assembled at runtime. In the planning-disabled setting, we remove planning instructions, reminders, plan injections, and the tool, while holding the execution loop, action interface, and context management fixed. Therefore, our results estimate the effect of this persistent planning scaffold rather than the effect of planning as a general reasoning strategy.

\subsection{Action space}
\label{sec:action}

The action space defines how the agents interact with the environment.
The predefined-tool provides \texttt{read\_file}, \texttt{write\_file}, \texttt{edit\_file}, \texttt{list\_files}, \texttt{glob\_files}, \texttt{grep\_text}, \texttt{web\_fetch}, and \texttt{bash}, as summarized in Table~\ref{tab:tools} (system prompt shown in Figure~\ref{prompt:tools_system}). Each tool has a typed argument schema and a description specifying its protocol, errors, and side effects; Appendix~\ref{app:tools} summarizes arguments and read-only status.
We exclude web search because SWE-Bench tasks originate from public GitHub issues, and search could expose the corresponding pull request and ground-truth patch~\citep{cao2026qwen3coder}. The bash-only setting removes predefined file, search, and web tools, leaving \texttt{bash} for general environment interaction (Figure~\ref{prompt:tools_system_bash_only}). Auxiliary tools controlled by other harness components remain unchanged: with planning enabled in T4/128k, both conditions retain \texttt{update\_plan} and \texttt{recall\_event}. Appendices~\ref{app:prompts_system} and~\ref{app:tools} provide the remaining prompts and tool descriptions. The intervention also changes how workspace modifications are tracked and validated. The predefined file tools enforce read-before-write checks, update the harness file state, and trigger automatic diagnostics after supported edits. The comparison should therefore be interpreted as the effect of the complete action interface, including tool availability, interface instructions, state tracking, and validation support, rather than as the isolated effect of tool count or action granularity.

\begin{table}[t]
\centering
\vspace{-8pt}
\caption{Tools exposed by the harness. Read-only tools do not modify the workspace or harness state and may execute concurrently within a model turn. The bash-only action-space condition removes the predefined workspace tools while retaining \texttt{bash} and the auxiliary tools required by the enabled planning and context-management components. The table lists the principal arguments. Appendix~\ref{app:tools} reproduces the whole natural-language descriptions.}
\vspace{-8pt}
\label{tab:tools}
\resizebox{\textwidth}{!}{
\begin{tabular}{lccl}
\toprule
\textbf{Tool} & \textbf{Read-only} & \textbf{Principal arguments} & \textbf{Description} \\
\midrule
\multicolumn{4}{c}{\textit{File I/O}} \\
\midrule
\texttt{read\_file}   & \cmark & \texttt{path, offset, limit}       & Read a file, returned with line numbers \\
\texttt{write\_file}  &     \xmark       & \texttt{path, content, overwrite}  & Create a file or overwrite an existing one \\
\texttt{edit\_file}   &       \xmark     & \texttt{path, old\_text, new\_text, replace\_all} & Replace an exact string in a file \\
\addlinespace
\midrule
\multicolumn{4}{c}{\textit{Search}} \\
\midrule
\texttt{list\_files}  & \cmark & \texttt{path, recursive}           & List the contents of a directory \\
\texttt{glob\_files}  & \cmark & \texttt{pattern, path}             & Find files matching a glob pattern \\
\texttt{grep\_text}   & \cmark & \texttt{query, path, include}      & Search file contents by regular expression \\
\addlinespace
\midrule
\multicolumn{4}{c}{\textit{Execution}} \\
\midrule
\texttt{bash}         &      \xmark      & \texttt{command, timeout\_seconds, cwd}     & Execute a shell command \\
\addlinespace
\midrule
\multicolumn{4}{c}{\textit{Web}} \\
\midrule
\texttt{web\_fetch}   & \cmark & \texttt{url, format}               & Fetch a web page as text or markdown \\
\addlinespace
\midrule
\multicolumn{4}{c}{\textit{Planning}} \\
\midrule
\texttt{update\_plan} &     \xmark       & \texttt{plan}                      & Create or update the stored task plan \\
\addlinespace
\midrule
\multicolumn{4}{c}{\textit{Context management}} \\
\midrule
\texttt{recall\_event} & \cmark & \texttt{id} & Return the verbatim content of a stored event (T2 and T4 only) \\
\bottomrule
\vspace{-8pt}
\end{tabular}}
\end{table}

\subsection{Context management}
\label{sec:context}

Context management determines how the growing interaction history is represented within a bounded context window. Existing methods are lossy or lossless. Lossy methods such as elision and summarization reduce context but may remove information that becomes useful later~\citep{xiao2024efficient,jiang2023llmlingua,wu2021recursively}, whereas lossless methods preserve recoverability through external storage and retrieval but require additional machinery and rely on the model to retrieve the right information~\citep{packer2023memgpt,park2023generative,ehrlich2026lcm,xu2026llm}. Our harness draws on both families through three composable mechanisms. Elision (M1) replaces the body of a stale tool observation with a short stub. Recall (M2) stores elided observations in the file system and exposes a \texttt{recall\_event} tool to read them back on demand, making elision reversible. Summarization (M3) folds older messages into a running natural-language summary. The summary is produced by a separate, tool-free call to the same model under evaluation, using the prompt in Appendix~\ref{app:prompts_context}. Elision reclaims tokens cheaply but discards detail, recall recovers that detail when needed, and summarization compresses history too old to keep verbatim. We combine them under two token thresholds: soft $B_1$ and hard $B_2$. The preamble (system prompt and initial task description) and a token-budgeted recent window of at least two turns remain verbatim; only the middle region is compacted. Once history exceeds $B_1$, the harness elides bulky tool observations in the middle region, storing originals externally and leaving stubs in their place (M1 and M2). If history still exceeds $B_2$, the harness summarizes the oldest middle events into the running summary (M3).
Algorithm~\ref{alg:context} gives the full procedure, and \texttt{recall\_event} remains available on every turn. Appendix~\ref{app:prompts_context} reproduces the summarization prompt and the stub and summary fragments inserted into model input.

\begin{wraptable}{r}{0.33\textwidth}
\centering
\vspace{-5pt}
\caption{The five context-management tiers, each enabling a subset of elision (M1), recall (M2), and summarization (M3).}
\vspace{-8pt}
\label{tab:context_management}
\begin{tabular}{c|ccc}
\toprule
 & \textbf{M1} & \textbf{M2} & \textbf{M3} \\
\midrule
Tier 0 & \xmark & \xmark & \xmark\\
Tier 1 & \cmark & \xmark & \xmark\\
Tier 2 & \cmark & \cmark & \xmark\\
Tier 3 & \xmark & \xmark & \cmark\\
Tier 4 & \cmark & \cmark & \cmark\\
\bottomrule
\end{tabular}
\end{wraptable}
To isolate each mechanism's contribution, we define five policy variants, summarized in Table~\ref{tab:context_management}. Tier 4 is the full three-mechanism configuration described in Algorithm~\ref{alg:context}. Tier 0 disables context management; trajectories that outgrow the window terminate with an error. Tier 1 uses elision alone (M1), replacing stale tool-observation bodies with short stubs and discarding the original content. Tier 2 adds recall (M2): elided observations are stored externally and recoverable via \texttt{recall\_event}, making elision reversible. Tier 3 uses summarization alone (M3), folding the middle region into a running summary without elision. Because Tiers 1--3 each have only one action, they operate at the hard threshold $B_2$, whereas Tier 4 elides at $B_1$ and summarizes at $B_2$.

\begin{algorithm}[t]
\caption{Per-turn context management (Tier 4)}
\label{alg:context}
\begin{algorithmic}[1]
\Require history $H$; soft and hard thresholds $B_1 < B_2$
\State append the new think, action, and observation to $H$
\If{the model invoked $\textproc{recall\_event}(id)$ this turn}
    \State read observation $id$ from the external store back into $H$ \Comment{M2}
\EndIf
\State keep the preamble and a budget-sized recent window (at least the last two turns) verbatim; let $M$ be the middle region
\If{$\mathrm{tokens}(H) \ge B_1$}
    \For{each bulky tool observation in $M$}
        \State store the original in the external store \Comment{M2}
        \State replace its body with a stub \Comment{M1}
    \EndFor
    \If{$\mathrm{tokens}(H) \ge B_2$}
        \State summarize the oldest events in $M$ into a running summary \Comment{M3}
    \EndIf
\EndIf
\State \Return $H$
\end{algorithmic}
\end{algorithm}
\vspace{-10pt}


\subsection{Other components}
\label{sec:substrate}
Beyond the three components mentioned above, the harness includes several supporting components that we hold fixed across all ablations. We highlight the three that are most important below.

\vspace{-10pt}

\paragraph{Safety}
Every action that reads or modifies the workspace passes through three gates. A workspace guard resolves each path and rejects any that escapes the project root, including through symlinks. A read-before-write check refuses to edit or overwrite a file that has not been read in the current session, and detects external modification through a content hash. A permission layer then classifies each action as allow, ask, or deny. Tool errors are returned to the model as observations rather than raised, so a failed action never crashes the loop and the model can recover from it.

\vspace{-10pt}

\paragraph{Post-edit diagnostics}
After the agent edits or writes a Python file, the harness runs a fast, read-only check on it with ruff, pyflakes, or a syntax-only fallback, and appends the findings to the tool result. This surfaces syntax errors, undefined names, and unused imports immediately, so the model can fix them before spending a turn on the tests.

\paragraph{Stuck detection}

A turn-based agent can spin, reissuing the same failing action until it exhausts its step budget. The harness monitors the tool log for streaks of identical calls, that is, consecutive calls with the same tool name and arguments. When such a streak reaches a threshold, it injects a one-time reminder to change approach, and when a streak of identical failing calls keeps growing, it ends the run early rather than grinding to the budget limit. The reminder texts and the thresholds are given in Appendix~\ref{app:prompts_stuck}.

\section{Experiment}
\label{sec:experiments}
\subsection{Setup}
\paragraph{Models.}
In our experiments, we use three sizes of the Nemotron-3 family~\citep{blakeman2025nvidia} (30B, 120B, and 550B). We additionally include Mistral-Medium-3.5-128B~\citep{mistral2026medium} from a different model family, to test whether our findings generalize beyond a single family. We price tokens at OpenRouter\footnote{\href{https://openrouter.ai/}{https://openrouter.ai/}, accessed August 2026.}, per 1M input\,/\,output tokens: \$0.05\,/\,\$0.20 (Nemotron-3-30B), \$0.08\,/\,\$0.45 (Nemotron-3-120B), \$0.50\,/\,\$2.20 (Nemotron-3-550B), and \$1.50\,/\,\$7.50 (Mistral-Medium-3.5).

\paragraph{Benchmarks.}
We evaluate on two long-horizon coding benchmarks: SWE-Bench Verified~\citep{jimenez2024swebench}, comprising 500 human-verified real GitHub issues, and Terminal-Bench 2.1~\citep{merrill2026terminal}, comprising 89 end-to-end tasks in a command-line environment. On both, we report two metrics: the task success rate, the fraction of tasks the agent resolves, and the mean cost per task, priced as described above.

\paragraph{Implementation Details}
\begin{itemize}[leftmargin=*]
\item \textbf{Models and serving} Three Nemotron-3 models and Mistral-Medium-3.5-128B are served locally with SGLang in BF16 precision. Temperature is set to 0, and top-$p$ is 0.95. We cap the output at 16{,}384 tokens per turn.

\item \textbf{Harness configuration} The harness is built on LangGraph,\footnote{\href{https://www.langchain.com/langgraph}{https://www.langchain.com/langgraph}} with benchmarks driven through Harbor,\footnote{\href{https://www.harborframework.com/}{https://www.harborframework.com/}} which owns each task's container and verifier while the host agent acts on the container. Each task runs for at most 300 steps. For context management, soft and hard context thresholds are 0.6 and 0.85 of the usable window, with the verbatim recent window budgeted at 0.3 and floored at two turns. Tool results are truncated to 24k characters, and up to eight read-only tools may run in parallel per step. Stuck detection issues a reminder after five consecutive identical tool calls or five consecutive identical failing calls, and terminates after eight consecutive identical failing calls.


\end{itemize}

\paragraph{Ablation Settings.}
We evaluate T0--T4 under 32k, 64k, 96k, and 128k context-window budgets, yielding $20$ settings per model--benchmark pair. All use the predefined tool set with planning enabled. The T4/128k setting serves as the baseline for the remaining component ablations. One matched setting disables planning, and another replaces the predefined tool set with the bash-only interface, with all other components fixed. Planning and the action space are evaluated only under T4/128k. The 20 context-management settings and the two additional component ablations produce 22 settings per model--benchmark pair, for a total of $176$ experimental settings across four models and two benchmarks. For each benchmark, we define three comparison families: management strategy vs. T0, planning on vs. off, and full tool set vs. bash-only. Within each family, we compare success rates using two-sided exact McNemar tests on task-paired outcomes and apply the Benjamini--Hochberg procedure to control the false discovery rate at 0.05.

\subsection{Main Results}
\label{sec:results}

\tcbset{takeaway/.style={
  colback=zoomblue!5!white, colframe=zoomblue!35!white,
  boxrule=0.7pt, arc=3pt, boxsep=1pt,
  left=6pt, right=6pt, top=4pt, bottom=4pt,
}}


\begin{table}[t]
    \centering
    \vspace{-10pt}
    \footnotesize
    \setlength{\tabcolsep}{4pt}
    \caption{Main results on \textbf{SWE-Bench Verified}, reporting success rate (SR, \%) and mean cost per task (\$). Each row pairs a context-window budget with a context-management tier: T0 applies no context management, T1 elision alone, T2 elision and recall, T3 summarization alone, and T4 all three mechanisms. The final two rows ablate a single component at the 128k/T4 setting: $-$plan disables planning, and bash replaces the structured tool set with a bare shell. \textbf{Bold} marks each model's highest SR. * denotes a significant difference from the matched baseline under a two-sided exact McNemar test with Benjamini--Hochberg-adjusted $q<0.05$.}
    \vspace{-8pt}
    \label{tab:results_swe}
    \resizebox{\textwidth}{!}{
    \begin{tabular}{@{}lccccccccc@{}}
    \toprule
    & & \multicolumn{2}{c}{Nemotron-3 30B} & \multicolumn{2}{c}{Nemotron-3 120B} & \multicolumn{2}{c}{Nemotron-3 550B} & \multicolumn{2}{c}{Mistral-3.5-128B} \\
    \cmidrule(lr){3-4}\cmidrule(lr){5-6}\cmidrule(lr){7-8}\cmidrule(lr){9-10}
    \multicolumn{2}{c}{Setting}  & SR (\%) & Cost (\$) & SR (\%) & Cost (\$) & SR (\%) & Cost (\$) & SR (\%) & Cost (\$) \\
    \midrule
    \multirow{5}{*}{32K} & T0 & \hphantom{0}9.40 & 0.04 & 11.40 & 0.05 & \hphantom{0}6.40 & 0.26 & 12.60 & 0.87 \\
     & T1 & 20.60\textsuperscript{*} & 0.09 & 43.00\textsuperscript{*} & 0.35 & 51.40\textsuperscript{*} & 2.57 & 64.40\textsuperscript{*} & 2.16 \\
     & T2 & 20.80\textsuperscript{*} & 0.09 & 42.20\textsuperscript{*} & 0.36 & 53.60\textsuperscript{*} & 2.70 & 66.20\textsuperscript{*} & 2.12 \\
     & T3 & 23.80\textsuperscript{*} & 0.11 & 39.60\textsuperscript{*} & 0.18 & 58.40\textsuperscript{*} & 1.25 & 63.20\textsuperscript{*} & 2.52 \\
     & T4 & 21.20\textsuperscript{*} & 0.11 & 42.20\textsuperscript{*} & 0.17 & 55.60\textsuperscript{*} & 1.45 & 63.80\textsuperscript{*} & 2.04 \\
    \midrule
    \multirow{5}{*}{64K} & T0 & 20.80 & 0.07 & 34.00 & 0.10 & 29.40 & 0.97 & 52.80 & 2.47 \\
     & T1 & 23.40 & 0.09 & 45.40\textsuperscript{*} & 0.39 & 63.80\textsuperscript{*} & 2.22 & \textbf{69.00}\textsuperscript{*} & 2.65 \\
     & T2 & 24.40 & 0.09 & 43.20\textsuperscript{*} & 0.32 & 64.60\textsuperscript{*} & 2.17 & 67.80\textsuperscript{*} & 2.61 \\
     & T3 & \textbf{26.40}\textsuperscript{*} & 0.10 & 43.60\textsuperscript{*} & 0.23 & 65.80\textsuperscript{*} & 1.78 & 68.40\textsuperscript{*} & 2.66 \\
     & T4 & 25.80\textsuperscript{*} & 0.08 & 41.80\textsuperscript{*} & 0.21 & 63.40\textsuperscript{*} & 1.78 & 66.00\textsuperscript{*} & 2.48 \\
    \midrule
    \multirow{5}{*}{96K} & T0 & 24.00 & 0.09 & 39.60 & 0.20 & 51.00 & 1.60 & 66.20 & 3.01 \\
     & T1 & 23.20 & 0.09 & 43.20 & 0.33 & 65.00\textsuperscript{*} & 2.25 & 68.80 & 3.13 \\
     & T2 & 23.60 & 0.09 & 45.40\textsuperscript{*} & 0.37 & 67.20\textsuperscript{*} & 2.33 & 66.20 & 3.15 \\
     & T3 & 23.60 & 0.09 & 46.20\textsuperscript{*} & 0.36 & 66.80\textsuperscript{*} & 2.32 & 67.60 & 3.12 \\
     & T4 & 24.80 & 0.09 & 43.40 & 0.30 & 66.80\textsuperscript{*} & 1.97 & 68.60 & 2.85 \\
    \midrule
    \multirow{7}{*}{128K} & T0 & 24.80 & 0.09 & 40.20 & 0.25 & 59.80 & 2.08 & 67.40 & 3.27 \\
     & T1 & 25.00 & 0.10 & 44.40 & 0.39 & 65.20\textsuperscript{*} & 2.47 & 68.60 & 3.25 \\
     & T2 & 26.00 & 0.10 & 45.20\textsuperscript{*} & 0.35 & 67.40\textsuperscript{*} & 2.64 & 67.00 & 3.10 \\
     & T3 & 23.60 & 0.11 & 44.00 & 0.35 & 65.80\textsuperscript{*} & 2.54 & 66.60 & 3.26 \\
     & T4 & 25.20 & 0.09 & 44.00 & 0.34 & 65.80\textsuperscript{*} & 2.33 & 68.60 & 3.14 \\
    \cmidrule(l{2pt}){2-10}
     & T4 w/o plan & 13.60\textsuperscript{*} & 0.02 & \textbf{46.60} & 0.25 & 67.80 & 3.31 & \textbf{69.00} & 4.65 \\
     & T4 bash only & 10.20\textsuperscript{*} & 0.03 & 42.40 & 0.35 & \textbf{69.40}\textsuperscript{*} & 1.11 & 45.40\textsuperscript{*} & 1.72 \\
    \bottomrule
    \vspace{-15pt}
    \end{tabular}}
    \end{table}

\begin{table}[t]
    \centering
    \vspace{-10pt}
    \footnotesize
    \setlength{\tabcolsep}{4pt}
    \caption{Main results on \textbf{Terminal-Bench 2.1}, reporting success rate (SR, \%) and mean cost per task (\$). Each row pairs a context-window budget with a context-management tier: T0 applies no context management, T1 elision alone, T2 elision and recall, T3 summarization alone, and T4 all three mechanisms. The final two rows ablate a single component at the 128k/T4 setting: $-$plan disables planning, and bash replaces the structured tool set with a bare shell. \textbf{Bold} marks each model's highest SR. * denotes a significant difference from the matched baseline under a two-sided exact McNemar test with Benjamini--Hochberg-adjusted $q<0.05$.}
    \vspace{-8pt}
    \label{tab:results_tb}
    \resizebox{\textwidth}{!}{
    \begin{tabular}{@{}lccccccccc@{}}
    \toprule
    & & \multicolumn{2}{c}{Nemotron-3 30B} & \multicolumn{2}{c}{Nemotron-3 120B} & \multicolumn{2}{c}{Nemotron-3 550B} & \multicolumn{2}{c}{Mistral-3.5-128B} \\
    \cmidrule(lr){3-4}\cmidrule(lr){5-6}\cmidrule(lr){7-8}\cmidrule(lr){9-10}
    \multicolumn{2}{c}{Setting}  & SR (\%) & Cost (\$) & SR (\%) & Cost (\$) & SR (\%) & Cost (\$) & SR (\%) & Cost (\$) \\
    \midrule
    \multirow{5}{*}{32K} & T0 & \hphantom{0}6.74 & 0.04 & 19.10 & 0.09 & 28.09 & 0.26 & 21.35 & 0.76 \\
     & T1 & 11.24 & 0.12 & \textbf{33.71}\textsuperscript{*} & 0.44 & 33.33 & 1.97 & 39.33\textsuperscript{*} & 2.15 \\
     & T2 & \hphantom{0}7.87 & 0.12 & 25.84 & 0.46 & 33.33 & 1.80 & 35.96\textsuperscript{*} & 2.45 \\
     & T3 & 14.61 & 0.11 & 22.47 & 0.13 & 38.20\textsuperscript{*} & 0.74 & 42.70\textsuperscript{*} & 1.91 \\
     & T4 & \textbf{17.98}\textsuperscript{*} & 0.11 & 21.35 & 0.14 & 32.58 & 0.82 & 42.70\textsuperscript{*} & 1.88 \\
    \midrule
    \multirow{5}{*}{64K} & T0 & 11.24 & 0.08 & 20.22 & 0.13 & 30.34 & 0.66 & 30.34 & 1.78 \\
     & T1 & 14.61 & 0.12 & 29.21 & 0.28 & 41.57 & 2.11 & 37.08\textsuperscript{*} & 2.74 \\
     & T2 & 11.24 & 0.12 & 26.97 & 0.32 & 40.45 & 2.35 & 37.08 & 2.17 \\
     & T3 & 14.61 & 0.12 & 26.97 & 0.20 & 43.82\textsuperscript{*} & 1.61 & 42.70\textsuperscript{*} & 2.51 \\
     & T4 & 13.48 & 0.10 & 25.84 & 0.22 & 44.94\textsuperscript{*} & 1.16 & 38.20 & 2.51 \\
    \midrule
    \multirow{5}{*}{96K} & T0 & 12.36 & 0.11 & 19.10 & 0.28 & 33.71 & 1.14 & 33.71 & 2.60 \\
     & T1 & 15.73 & 0.14 & 22.47 & 0.23 & 42.70 & 2.47 & 37.08 & 3.00 \\
     & T2 & 11.24 & 0.15 & 21.35 & 0.37 & 43.82\textsuperscript{*} & 2.71 & 38.20 & 3.12 \\
     & T3 & \textbf{17.98} & 0.16 & 28.09 & 0.32 & 34.83 & 2.14 & 39.33 & 2.84 \\
     & T4 & 13.48 & 0.13 & 25.84 & 0.25 & 44.94\textsuperscript{*} & 1.66 & 35.96 & 2.79 \\
    \midrule
    \multirow{7}{*}{128K} & T0 & 11.24 & 0.12 & 25.84 & 0.27 & 34.83 & 1.65 & 34.83 & 3.32 \\
     & T1 & 10.11 & 0.15 & 22.47 & 0.40 & 40.45 & 2.68 & 40.45 & 3.71 \\
     & T2 & 16.85 & 0.15 & 25.84 & 0.32 & 40.45 & 2.37 & 37.08 & 4.16 \\
     & T3 & 12.36 & 0.16 & 26.97 & 0.30 & 37.08 & 2.26 & 38.20 & 3.65 \\
     & T4 & 13.48 & 0.14 & 28.09 & 0.28 & 44.94\textsuperscript{*} & 2.43 & 37.08 & 2.22 \\
    \cmidrule(l{2pt}){2-10}
     & T4 w/o plan & \hphantom{0}8.99 & 0.08 & 28.09 & 0.38 & 46.07 & 2.52 & 39.33 & 3.71 \\
     & T4 bash only & \hphantom{0}3.37\textsuperscript{*} & 0.02 & 23.56 & 0.41 & \textbf{50.56} & 1.70 & \textbf{43.82} & 2.75 \\
    \bottomrule
    \vspace{-10pt}
    \end{tabular}}
    \end{table}


\begin{figure}[t]
\centering
\vspace{-8pt}
\includegraphics[width=\textwidth]{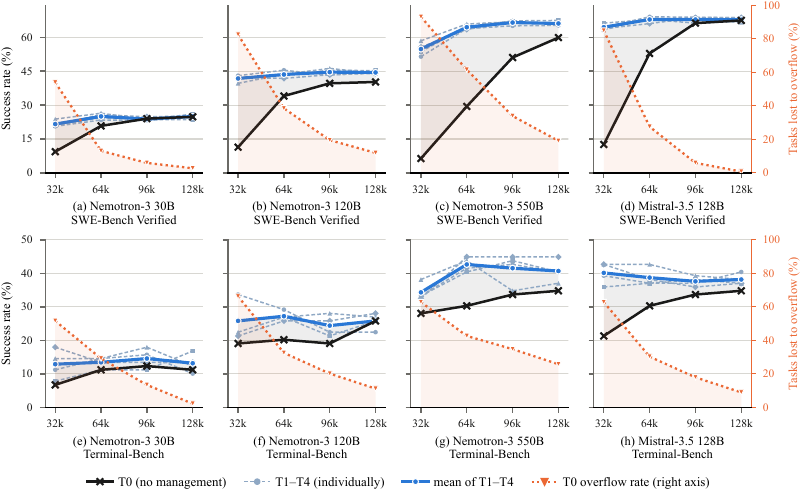}
\vspace{-8pt}
\caption{\textbf{Success rate and window-overflow rate against the context-window budget.} Left axis: success rate; right axis (\textcolor{orange}{orange}): the fraction of tasks T0 loses to window overflow. In each panel, the black curve shows T0, the light-blue curves show T1--T4 individually, and the solid blue curve shows their mean. Every managed tier overflows on exactly zero tasks at every budget, so a single overflow curve suffices.}
\label{fig:window}
\vspace{-12pt}
\end{figure}

\paragraph{The value of context management grows as the context-window budget shrinks.}
For each model and context-window budget, we define the value of context management as the success-rate gap between the managed tiers (T1--T4) and no management (T0). Averaged across the models, the managed--T0 gap shrinks steadily across 32k, 64k, 96k, and 128k windows: from $35.7$ to $15.9$, $5.5$, and $2.7$ percentage points on SWE-Bench, and from $9.5$ to $7.5$, $4.8$, and $2.8$ on Terminal-Bench (Tables~\ref{tab:results_swe} and~\ref{tab:results_tb}). Figure~\ref{fig:window} shows that the narrowing managed--T0 gap tracks the decline in T0 window-overflow failures as the window increases. Across these budgets, the model-averaged T0 overflow rate falls from $78.7\%$ to $8.7\%$ on SWE-Bench and from $61.0\%$ to $12.1\%$ on Terminal-Bench, while all managed tiers have zero overflow failures throughout. Thus, context management is valuable largely because it prevents premature truncation when the window binds; as more unmanaged trajectories fit within the window, its marginal accuracy benefit shrinks and becomes more model-dependent.

\begin{tcolorbox}[takeaway]
\textbf{Takeaway:} Context management reduces the sensitivity of task success to context-window capacity, enabling effective execution under tighter context budgets.
\end{tcolorbox}

\begin{figure}[t]
\centering
\vspace{-10pt}
\includegraphics[width=\textwidth]{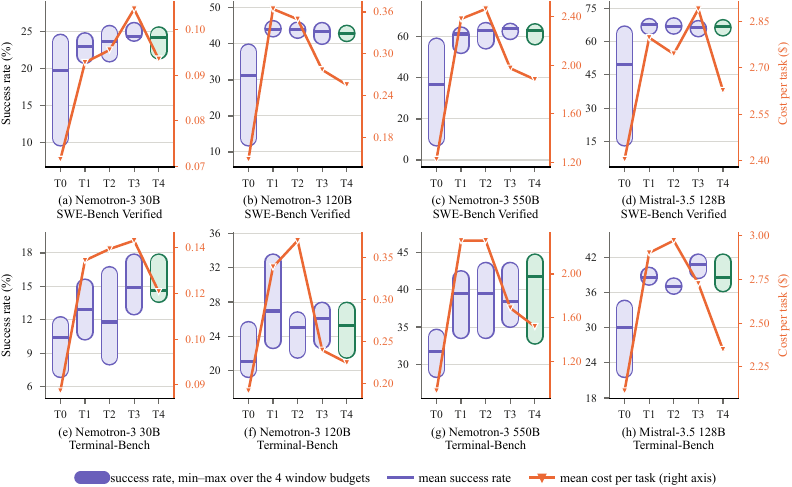}
\vspace{-10pt}
\caption{\textbf{Success rate and cost across context-management tiers.} For each tier, the capsule on the left axis spans the minimum and maximum success rates over the 32k, 64k, 96k, and 128k context-window budgets, and the horizontal bar marks their mean; capsule height therefore indicates sensitivity to the window budget. T4, the default tier, is highlighted in \textcolor[HTML]{1F7A54}{green}. The \textcolor[HTML]{EB6834}{orange} line is read against the right axis and reports the mean cost per task over the same four budgets.}
\vspace{-10pt}
\label{fig:tiercost}
\end{figure}

\begin{figure}[t]
\centering
\includegraphics[width=\textwidth]{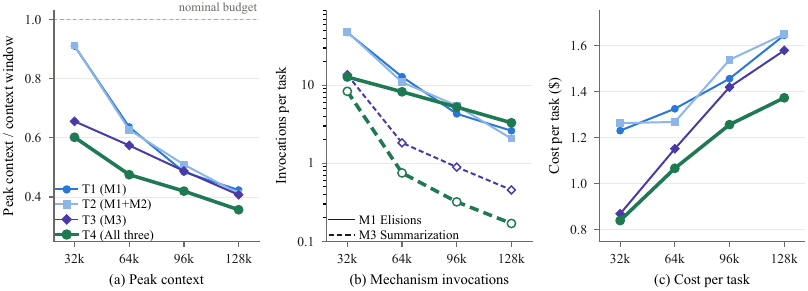}
\vspace{-10pt}
\caption{\textbf{Context compression, mechanism use, and cost across context-window budgets.} Each curve reports the equal-weight mean over four models and two benchmarks. \textbf{(a)} Mean peak context divided by the corresponding nominal context-window budget; lower values indicate more effective context compression. \textbf{(b)} Mean invocations per task for elision (M1) and summarization (M3), shown on a logarithmic scale. Solid curves show M1 invocations for T1, T2, and T4, while dashed curves show M3 invocations for T3 and T4, the only tiers that enable summarization. \textbf{(c)} Mean cost per task in dollars (\$).}
\vspace{-10pt}
\label{fig:tiermechanism}
\end{figure}

\paragraph{T4 offers the best accuracy--cost trade-off among context-management strategies.}
Figure~\ref{fig:tiercost} shows that T4 achieves success rates comparable to T1--T3, with the lowest cost in seven of eight model--benchmark panels. To explain this cost profile, we normalize mean peak context by the corresponding nominal context-window budget. Figure~\ref{fig:tiermechanism}(a) shows that at 32k, T1 and T2 trajectories still reach approximately the full window, whereas T3 and T4 keep peak context substantially below it. Across model--benchmark pairs, T4 has the lowest average peak-context ratio at all four window budgets. Figure~\ref{fig:tiermechanism}(b) compares M1 elisions and M3 summarization calls. T4 invokes M1 less often than T1 and T2 at 32k and 64k and at comparably low rates at larger windows; it also invokes M3 less often than T3 on average at every budget. Averaged across the eight model--benchmark pairs, T4 has the lowest mean cost per task at every window budget (Figure~\ref{fig:tiermechanism}(c)). These results suggest that T4's early elision handles many cases before summarization is needed, reducing costly LLM summarization calls and helping explain its lower cost.

\begin{tcolorbox}[takeaway]
\textbf{Takeaway:} With comparable accuracy across context-management strategies, T4 achieves the best overall cost profile by using cheap early elision to reduce reliance on LLM summarization.
\end{tcolorbox}


\paragraph{Recall (M2) is rarely used and does not improve accuracy over elision alone.}
\label{sec:recall_usage}
Recall (M2) makes elision reversible by storing elided observations externally and exposing \texttt{recall\_event} for on-demand retrieval. Because T1 and T2 differ only in M2 availability, they provide a matched comparison of its value.
Across 32 model--benchmark--window comparisons, T2 outperforms T1 in 15 settings, underperforms in 14, and ties in three; the equal-weight mean difference is $-0.36$ percentage points ($+0.40$ on SWE-Bench, $-1.12$ on Terminal-Bench).
Table~\ref{tab:recall_usage} shows that recall is rarely invoked.
Among the 64 T2 and T4 settings, 36 ($56.3\%$) never call \texttt{recall\_event}, the median invocation rate is zero, and the equal-weight mean falls from $0.540$ calls per task at 32k to $0.069$, $0.011$, and $0.007$ at 64k, 96k, and 128k; the 16 planning and action-space settings at T4/128k record no recall calls at all.
Recall use is therefore concentrated under the greatest context pressure and almost entirely in Nemotron-3 30B; Nemotron-3 550B and Mistral-Medium-3.5-128B rarely invoke it.
Even the heaviest-use configuration, Nemotron-3 30B on Terminal-Bench at 32k under T2, averages $4.326$ calls per task and scores $3.37\%$ below T1. 
These results suggest that lossless storage adds machinery most models seldom use, and retrieving elided observations does not consistently translate into completed tasks. 


\begin{tcolorbox}[takeaway]
\textbf{Takeaway:} Models, especially stronger ones, almost never call \texttt{recall\_event} to recover elided observations, so lossless recall yields no accuracy gain over elision alone.
\end{tcolorbox}

\begin{figure}[t]
\centering
\includegraphics[width=\textwidth]{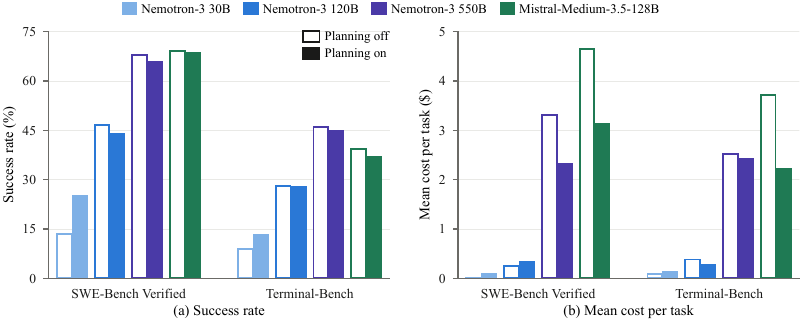}
\vspace{-10pt}
\caption{\textbf{Absolute performance and cost with and without planning.} Bars use T4 context management, a 128k context-window budget, and the full tool set. Within each model pair, the hollow left bar disables planning and the filled right bar enables it; this encoding applies to both panels. \textbf{(a)} reports success rate, and \textbf{(b)} reports mean cost per task.}
\label{fig:planningeffect}
\vspace{-10pt}
\end{figure}

\paragraph{Planning improves success rate for the weaker model and saves cost for the stronger models.}
We compare planning on and off under T4/128k with the full tool set (Figure~\ref{fig:planningeffect}). For Nemotron-3 30B, planning increases success rate by $11.6$ percentage points on SWE-Bench and $4.5$ points on Terminal-Bench, at higher cost on both benchmarks. For Nemotron-3 120B, planning yields no consistent success-rate gain; it increases cost on SWE-Bench but reduces it on Terminal-Bench. For Nemotron-3 550B and Mistral-Medium-3.5-128B, planning reduces cost on both benchmarks, accompanied by small decreases in success rate. On SWE-Bench, their costs fall by approximately $30\%$ and $32\%$, respectively, while success rates decrease by $2.0$ and $0.4$ percentage points (Tables~\ref{tab:results_swe} and~\ref{tab:results_tb}).
The execution statistics help explain these cost changes (Table~\ref{tab:planning_behavior}). Planning increases turns, tool calls, and average input tokens per call for Nemotron-3 30B on both benchmarks. All three measures decrease for Mistral on both benchmarks and for Nemotron-3 550B on SWE-Bench; the changes for Nemotron-3 550B on Terminal-Bench are smaller. Section~\ref{sec:analysis} traces these changes to individual trajectories: planning extends Nemotron-3 30B runs that would otherwise terminate before an edit, while the removed turns for Nemotron-3 550B and Mistral-Medium-3.5-128B are largely post-edit verification.

\begin{tcolorbox}[takeaway]
\textbf{Takeaway:} Planning trades additional computation for accuracy on the weaker model, but primarily reduces cost on the stronger models; its value at intermediate capability remains task-type-dependent.
\end{tcolorbox}
\begin{table}[t]
    \centering
    \small
    \setlength{\tabcolsep}{4pt}
    \caption{Percentage change from planning off to planning on at T4, a 128k context-window budget, and the full tool set. Positive values indicate an increase with planning and negative values a decrease. Input tokens avg. is the mean number of input tokens per agent call.}
    \vspace{-8pt}
    \label{tab:planning_behavior}
    \begin{tabular*}{\textwidth}{@{\extracolsep{\fill}}lrrr@{\hspace{8pt}}rrr@{}}
        \toprule
        & \multicolumn{3}{c}{SWE-Bench Verified}
        & \multicolumn{3}{c}{Terminal-Bench} \\
        \cmidrule(r{6pt}){2-4}\cmidrule(l){5-7}
        Model
        & \# Turns & \# Tool Calls & \makecell{Input Tokens\\Avg.}
        & \# Turns & \# Tool Calls & \makecell{Input Tokens\\Avg.} \\
        \midrule
        Nemotron-3 30B  & $+293.2\%$ & $+474.0\%$ & $+104.1\%$ & $+66.7\%$ & $+83.8\%$ & $+24.0\%$ \\
        Nemotron-3 120B & $+20.1\%$  & $+42.7\%$  & $+10.3\%$  & $-15.0\%$ & $-23.1\%$ & $-12.0\%$ \\
        Nemotron-3 550B & $-24.3\%$  & $-24.5\%$  & $-12.4\%$  & $+6.8\%$  & $+7.9\%$  & $-1.7\%$ \\
        Mistral-Medium-3.5-128B
                         & $-23.2\%$  & $-23.3\%$  & $-15.1\%$  & $-22.5\%$ & $-21.9\%$ & $-13.0\%$ \\
        \bottomrule
    \end{tabular*}
\end{table}

\begin{figure}[t]
\centering
\includegraphics[width=\textwidth]{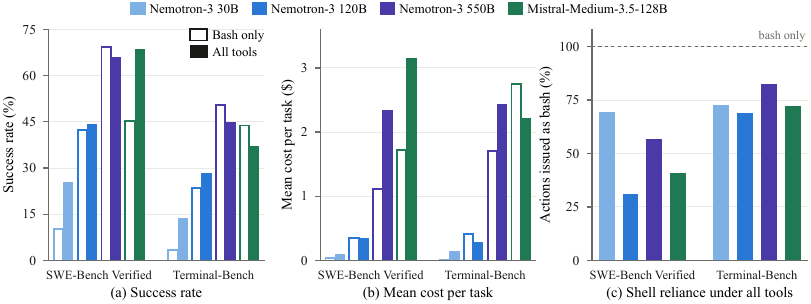}
\vspace{-12pt}
\caption{\textbf{Effect of the action space.} Results compare bash-only with the full tool set under T4 context management, a 128k context-window budget, and planning on. \textbf{(a)} reports success rate, \textbf{(b)} reports mean cost per task, and \textbf{(c)} reports the proportion of tool calls issued through \texttt{bash} when the full tool set is enabled.}
\label{fig:tooleffect}
\vspace{-12pt}
\end{figure}

\paragraph{The predefined tool set scaffolds weaker models, while bash-only can benefit stronger ones.}
Under the default T4/128k configuration with planning on, we compare the full tool set against bash-only. Figure~\ref{fig:tooleffect} shows that the predefined tool set benefits the weaker Nemotron-3 models most (Tables~\ref{tab:results_swe} and~\ref{tab:results_tb}).
For Nemotron-3 30B, the predefined tool set raises success by 15.0\% on SWE-Bench and 10.1\% on Terminal-Bench. These gains reflect alignment between the model's learned action vocabulary and the harness interface. Without predefined tools, the model falls back on tool-call patterns acquired during training rather than translating intended operations into \texttt{bash}. Because the bash-only registry lacks these tools, the harness cannot resolve emitted calls into executable actions. On Terminal-Bench, 66\% of bash-only trajectories terminate after such out-of-interface emissions, shortening the average trajectory from 71 to 15 turns. The predefined tool set scaffolds Nemotron-3 30B by exposing an action vocabulary it can invoke reliably.
For Nemotron-3 120B, accuracy gains shrink to $1.6\%$ on SWE-Bench and $4.5\%$ on Terminal-Bench but accompany more efficient execution. Unlike 30B, 120B does not benefit from longer trajectories: the predefined tool set shortens average runs from $101$ to $77$ turns on SWE-Bench and $96$ to $70$ on Terminal-Bench without increasing cost.
For Nemotron-3 550B, bash-only improves success rate by $3.6\%$ on SWE-Bench and $5.6\%$ on Terminal-Bench while reducing cost by $53\%$ and $30\%$, respectively. Bash-only trajectories issue $32\%$ fewer calls on SWE-Bench and $24\%$ fewer on Terminal-Bench. This is consistent with the model using denser, more composite shell commands rather than distributing work across predefined actions. For this model, the predefined tool set appears to add action-selection and interaction overhead rather than useful scaffolding. Section~\ref{sec:analysis} examines this shift at the edit level.
Mistral-Medium-3.5-128B exposes the workload boundary of this crossover. The full tool set improves success by $23.2\%$ on SWE-Bench, but bash-only adds $6.7\%$ on Terminal-Bench. Figure~\ref{fig:tooleffect}(c) explains the reversal: with all tools, Mistral issues $71.9\%$ of Terminal-Bench workspace actions via bash versus $40.4\%$ on SWE-Bench, while predefined-tool use falls from $31.4$ to $13.1$ calls per task. Terminal-Bench is therefore more shell-centric, so removing competing workspace tools better matches the model's preferred actions; on SWE-Bench, predefined read, search, and edit actions remain important.
Additional bash-only failures arise before repair: 32.8\% of Mistral's bash-only SWE-Bench runs end without editing a file, versus 1.2\% with the full tool set (Table~\ref{tab:app_termination_stage_full}), and among unresolved runs, the share never reaching the correct file rises from 16.0
The Mistral result is therefore an accuracy--cost trade-off rather than uniform dominance, showing that the preferred action space depends on both capability and task types.

\begin{tcolorbox}[takeaway]
\textbf{Takeaway:} The predefined tool set scaffolds weaker models that cannot reliably express intent through bash alone, but can add overhead for stronger models capable of composing complex, multi-step shell commands. The crossover also depends on how shell-centric the task types is.
\end{tcolorbox}

\begin{figure}[t]
\centering
\includegraphics[width=\textwidth]{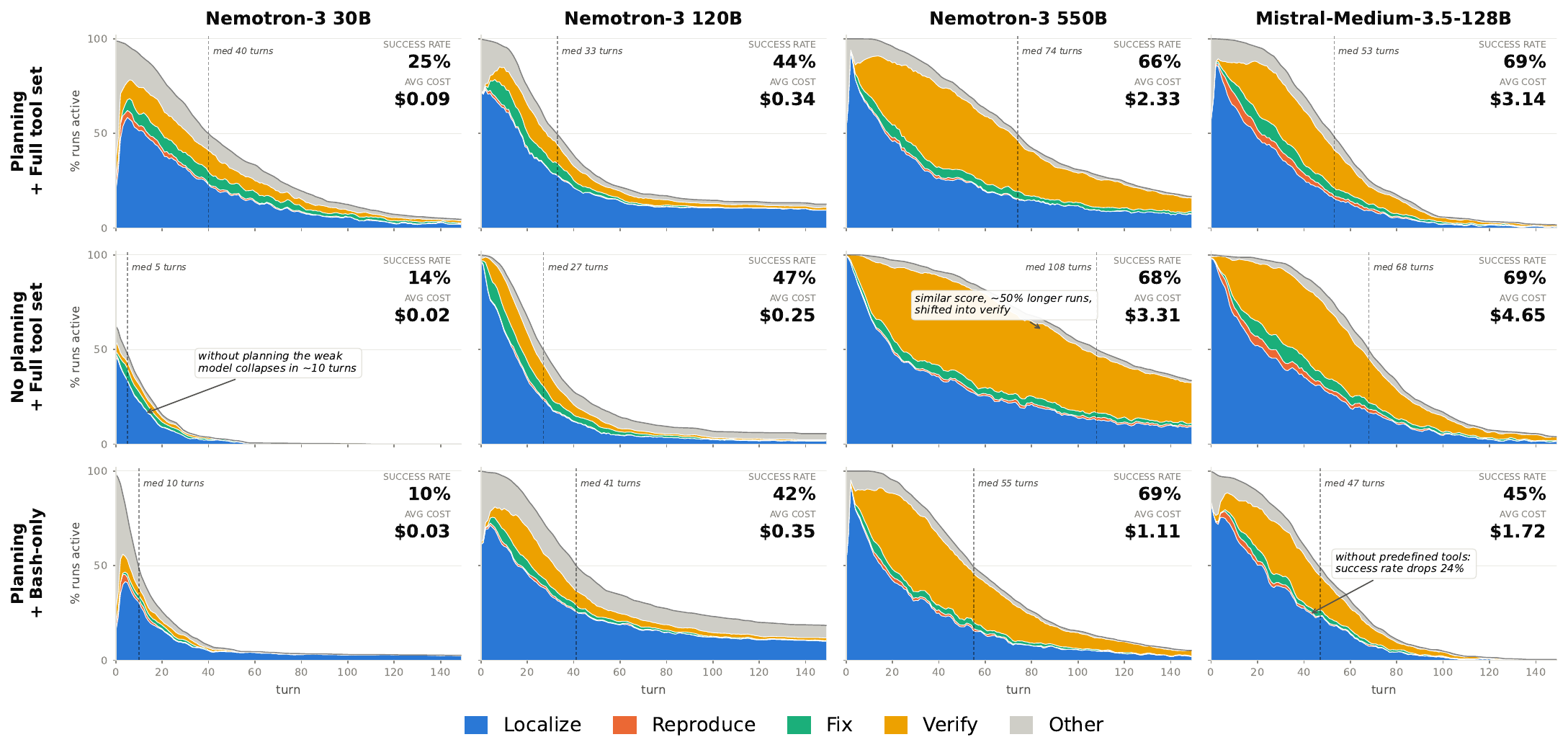}
\caption{\textbf{SWE-Bench trajectory profiles across harness configurations.} Columns correspond to models and rows to harness settings, all under T4 context management and a 128k context-window budget. The stacked areas show the fraction of runs still active at each turn, decomposed into Localize, Reproduce, Fix, Verify, and Other behavior. Dashed vertical lines mark the median trajectory length.}
\label{fig:swe-bench-behavior}
\end{figure}

\begin{figure}[t]
\centering
\includegraphics[width=\textwidth]{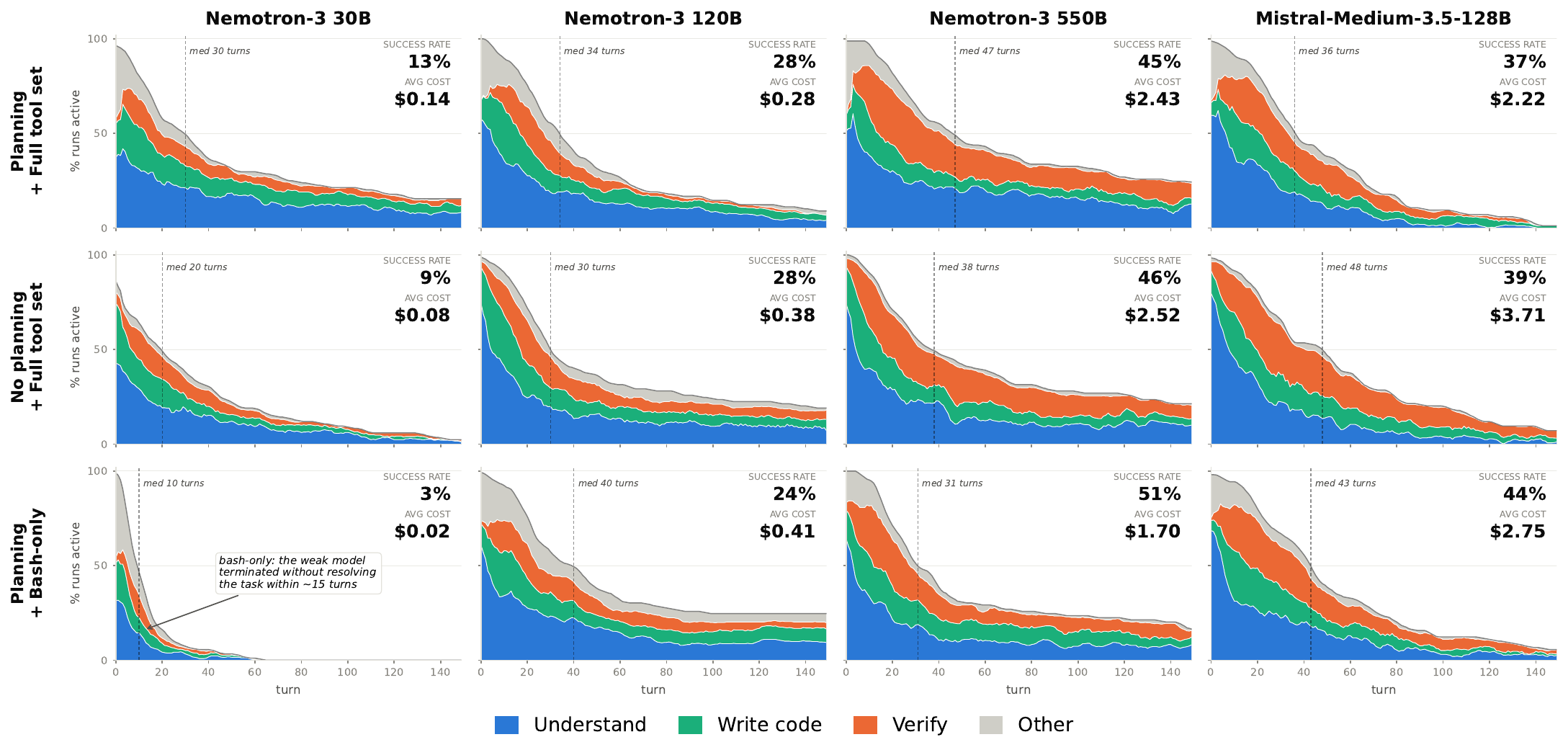}
\caption{\textbf{Terminal-Bench trajectory profiles across harness configurations.} Columns correspond to models and rows to harness settings, all under T4 context management and a 128k context-window budget. The stacked areas show the fraction of runs still active at each trajectory step, decomposed into Understand, Write Code, Verify, and Other behavior. Dashed vertical lines mark the median trajectory length.}
\label{fig:terminal_bench_behavior}
\end{figure}
\section{Analysis}
\label{sec:analysis}

In this section, we analyze the annotated trajectories to identify the behavioral mechanisms underlying the main results. While the aggregate results establish \emph{which} harness components matter, they do not explain \emph{how} these effects arise. Each turn in agent trajectories is labeled with its workflow phase by an LLM judge using the taxonomies in Appendix~\ref{app:traj_analysis}; the judge's agreement with human annotators is reported in Appendix~\ref{app:human_eval}. We first show that context management extends execution trajectories without substantially altering agent behavior, motivating its use at T4/128k throughout the remaining analysis. We then examine why planning lengthens trajectories for the weakest model but shortens them for the strongest ones, and how the bash-only interface changes the way actions are expressed.

\paragraph{Context management extends execution trajectories without substantially altering agent behavior.}
\begin{wraptable}{r}{0.4\textwidth}
\centering
\vspace{-10pt}
\small
\resizebox{\linewidth}{!}{%
\begin{tabular}{@{}lrrrr@{}}
\toprule
 & \multicolumn{2}{c}{W/o edit (\%)} & \multicolumn{2}{c}{At loc.\ (\%)} \\
\cmidrule(lr){2-3} \cmidrule(lr){4-5}
Model & Off & On & Off & On \\
\midrule
Nemotron-3 30B & 68.6 & 27.8 & 58.4 & 10.4 \\
Nemotron-3 120B & 15.6 & 24.4 & 11.0 & 17.8 \\
Nemotron-3 550B & 1.4 & 2.4 & 0.2 & 0.6 \\
Mistral-Medium-3.5-128B & 2.0 & 1.2 & 1.0 & 0.6 \\
\bottomrule
\end{tabular}%
}
\vspace{-8pt}
\caption{SWE-Bench runs terminated without an edit, and the subset stalled at localization (every labeled turn still in the Localize phase of the encoding in Appendix~\ref{app:traj_analysis}), with planning off and on (T4/128k, full tool set).}
\label{tab:termination_stage}
\vspace{-10pt}
\end{wraptable}
At 32k, the tiers differ primarily in trajectory length. Without context management, the proportion of active SWE-Bench runs declines sharply, with median lengths of 20--30 turns across the four models. Most runs terminate during the Localize phase, and few reach Verify (Figure~\ref{fig:swe_behavior_ctx32k}). All managed tiers extend execution while largely preserving phase ordering and proportions at comparable turns. Median lengths increase to approximately 50--180 turns, depending on the model and tier, allowing runs to progress to verification. Terminal-Bench exhibits a similar pattern (Figure~\ref{fig:tb_behavior_ctx32k}). At 128k, differences across tiers largely diminish. Median trajectory lengths range from 39--42 turns for Nemotron-3 30B and 70--74 turns for Nemotron-3 550B. Re-patch counts differ by fewer than two per task, while the proportion of runs terminating without an edit varies by less than four percentage points for three of the four models (Figures~\ref{fig:swe_behavior_ctx128k} and~\ref{fig:tb_behavior_ctx128k}; Tables~\ref{tab:app_actionspace_granularity_full} and~\ref{tab:app_termination_stage_full}). These results indicate that context management primarily extends execution trajectories without substantially altering agent behavior. We therefore fix context management at T4/128k in the remaining analyses to examine how planning and the action space affect trajectory structure.


\paragraph{Planning sustains the weakest model's trajectory long enough to attempt an edit.}
Figure~\ref{fig:swe-bench-behavior} helps explain the performance gain for Nemotron-3 30B. Disabling planning reduces the median SWE-Bench trajectory length from 40 to 5 turns. Without planning, 68.6\% of runs terminate without an edit and 58.4\% terminate during Localize, compared with 27.8\% and 10.4\%, respectively, with planning (Table~\ref{tab:termination_stage}). These results suggest that planning helps the least capable model sustain execution through an initial edit attempt, with the additional turns contributing to higher cost. For Nemotron-3 120B, trajectory-length distributions nearly overlap, consistent with the absence of an accuracy gain.

\paragraph{Planning shortens SWE-Bench trajectories for stronger models by reducing post-edit verification.}
Figure~\ref{fig:swe-bench-behavior} shows that planning reduces the median trajectory length from 108 to 74 turns for Nemotron-3 550B and from 68 to 53 turns for Mistral-Medium-3.5-128B. The phase composition attributes most of this reduction to verification rather than localization or fixing. Consistently, planning produces little change in the fraction of runs terminating without an edit or failing at file localization (see Tables~\ref{tab:app_termination_stage_full} and~\ref{tab:app_failure_stage}), while substantially reducing turns and tool calls (see Table~\ref{tab:app_actionspace_granularity_full}). These results suggest that planning primarily improves stopping behavior by reducing redundant post-edit verification, rather than accelerating localization or repair.

\paragraph{Planning's cost effect on Terminal-Bench depends on how it reshapes the trajectory-length distribution.}
Figure~\ref{fig:terminal_bench_behavior} shows that planning has two opposing effects: it extends trajectories that would otherwise terminate prematurely, while truncating the tail of excessively long runs. The balance between these effects determines the model-specific cost changes in Figure~\ref{fig:planningeffect}(b). For Nemotron-3 120B, planning truncates the long-running tail and reduces cost by roughly 26\%. The effect is smaller for Nemotron-3 550B (3.6\%), whose exploratory tail is less compressed, but larger for Mistral-Medium-3.5-128B (about 40\%), where both medium- and long-running trajectories shorten. Nemotron-3 30B shows the opposite pattern: by preventing premature termination, planning increases cost by 75\%.

\begin{wraptable}{r}{0.5\textwidth}
\centering
\caption{Action granularity with the full tool set vs.\ bash-only (planning on, T4/128k): mean re-patches per task (edits to an already-edited file) and median largest edit on SWE-Bench, and the create/replace share of file-writing actions on Terminal-Bench.}
\label{tab:actionspace_granularity}
\small
\resizebox{\linewidth}{!}{%
\begin{tabular}{@{}lrrrrrr@{}}
\toprule
 & \multicolumn{2}{c}{Re-patch} & \multicolumn{2}{c}{Edit lines} & \multicolumn{2}{c}{Create (\%)} \\
\cmidrule(lr){2-3} \cmidrule(lr){4-5} \cmidrule(lr){6-7}
Model & Tools & Bash & Tools & Bash & Tools & Bash \\
\midrule
Nemotron-3 30B & 3.3 & 0.4 & 26 & 13 & 28 & 64 \\
Nemotron-3 120B & 2.8 & 2.2 & 22 & 24 & 39 & 76 \\
Nemotron-3 550B & 4.6 & 1.5 & 18 & 54 & 51 & 76 \\
Mistral-Medium-3.5-128B & 3.0 & 1.3 & 87 & 68 & 28 & 57 \\
\bottomrule
\end{tabular}%
}
\end{wraptable}
\paragraph{Bash-only enables larger code-writing actions and fewer interactions.}
Figure~\ref{fig:terminal_bench_behavior} shows this most clearly for Nemotron-3 550B on Terminal-Bench: switching from the full tool set to bash-only shortens the median trajectory from 47 to 31 actions while raising the share of Write-code actions from 16\% to 27\%, so a larger fraction of a shorter trajectory is spent producing code. This pattern suggests that the shell lets capable models bundle several low-level operations into one command or script, whereas the structured interface spreads the same work across many smaller interactions. Table~\ref{tab:actionspace_granularity} provides the corresponding fine-grained evidence. Across all four models, bash-only reduces repeated patching of already edited files, from 3.3 to 0.4 re-patches for Nemotron-3 30B, 2.8 to 2.2 for 120B, 4.6 to 1.5 for 550B, and 3.0 to 1.3 for Mistral. On Terminal-Bench, bash-only also shifts file-writing toward coarser create-or-replace actions, increasing their share from 28\% to 64\% for 30B, 39\% to 76\% for 120B, 51\% to 76\% for 550B, and 28\% to 57\% for Mistral. Thus, predefined tools lower the complexity of each individual action, but they often do so by increasing the number of interactions and incremental repair cycles needed to express the same operation.

\section{Related Work}

\subsection{Coding Agents}
The engine of a coding agent is a capable, increasingly code- and agent-specialized language model. Recent frontier models, including Qwen3-Coder~\citep{cao2026qwen3coder}, Kimi~K2.5 and K2.6~\citep{team2026kimi25,team2026kimik26}, GLM-5~\citep{zeng2026glm}, and Mistral~Medium~3.5~\citep{mistral2026medium}, are designed or evaluated for code generation and agentic use. Their progress is measured by execution-based benchmarks such as repository-level issue resolution in SWE-Bench~\citep{jimenez2024swebench} and end-to-end command-line task completion in Terminal-Bench~\citep{merrill2026terminal}. Because these benchmarks evaluate complete agent systems, their scores conflate model capability with the control loop, action interface, and context-management policy. This motivates controlled comparisons that hold the agent substrate fixed while varying one component.

\subsection{Coding Harness}
A coding harness is the software layer that turns LLMs into agents, comprising the control loop, tool interface, and context management through which they act on a codebase. Such harnesses now drive coding-agent products and frameworks like Claude Code~\citep{anthropic2025claudecode}, Codex~\citep{openai2025codex}, OpenCode~\citep{sst2025opencode}, and OpenHands~\citep{wang2025openhands}. Research on their design has shown that the harness, not the model alone, governs performance~\citep{yang2024sweagent}. Later systems add long-term memory for the harness~\citep{wong2025confucius} or add task graphs and modular sub-agents~\citep{chen2024coder,arora2024masai}, and recent surveys catalogue the design space~\citep{rombaut2026scaffold}. These systems generally optimize bundled designs. Although several report local ablations, few compare the same implementation-level interventions across model capability and context-window budgets.

A growing body of work therefore ablates coding harnesses to determine which components matter. At the system level, constrained localize, repair, and validate pipelines such as Agentless~\citep{xia2024agentless} and AutoCodeRover~\citep{zhang2024autocoderover} show that strong performance can be achieved without highly elaborate agent architectures. Finer-grained studies further indicate that the preferred design depends on the model. AgentArch~\citep{bogavelli2025agentarch} reports model-specific architecture preferences, while a large trajectory study finds that behavioral differences between coding-agent frameworks change across model generations~\citep{mehtiyev2026beyond}. Individual interventions can likewise become less useful or even harmful for stronger models, as observed for prompt-engineering strategies in software engineering~\citep{wang2024advanced}. Closest to our work, \citet{liu2026more} study prompt-level scaffolding on short-horizon reasoning tasks using a full-factorial design, where context-window pressure is not explicitly examined. We instead study implementation-level planning, workspace action interfaces, and context-management policies over substantially longer coding trajectories, where the context window can become a binding constraint, and explicitly manipulate the available context-window budget. This setting allows us to characterize how the conditional effects of these harness components vary with both model capability and resource availability.

\subsection{Context Management}
As the software-engineering tasks grow more challenging, an agent needs more steps to solve them, which drives its interaction history beyond the model's context window. Therefore, many studies have proposed context management strategies to balance task performance with context length. First, static methods apply fixed, hand-designed strategies, such as eliding or truncating stale content~\citep{xiao2024efficient,jiang2023llmlingua}, retrieving earlier content on demand from an external store~\citep{packer2023memgpt,park2023generative}, and summarizing history into a compact state~\citep{wu2021recursively}. Second, dynamic methods train the model, often via reinforcement learning, to manage its own context~\citep{wu2025resum,kang2025acon,li2025sculptor,zhang2026memory}. Such mechanisms are typically introduced as improvements and evaluated on a single model, which leaves their central dependence uncharacterized. The value of managing context cannot be separated from how much window there is to manage against, nor from whether the model is strong enough to be helped rather than starved by compaction. We characterize this dependence precisely, sweeping the context window from 32k to 128k tokens across two model families and three model sizes.

\section{Conclusion}

We present a controlled empirical study that estimates the conditional effects of three central coding-harness components: context management, planning, and the action space, across four models and two long-horizon coding benchmarks. Context management matters most under tight context-window budgets, and among its policies T4 achieves the lowest aggregate cost at broadly similar success rates by applying rule-based elision before selective LLM summarization. Planning improves success at additional cost for weaker models but mainly reduces cost, with small decreases in success rate, for stronger models. Predefined tools raise success rates for models with weak bash control, whereas bash-only yields higher success at lower cost for bash-capable models, most clearly on shell-centric task types. Trajectory analysis ties each effect to a distinct behavioral change and, through it, to the limitation the component addresses: context management extends execution trajectories without substantially altering agent behavior and is most beneficial under tight context budgets; planning sustains the trajectories of models that abandon tasks too early and trims repeated verification in models that verify too long; and structured tools support models with limited shell proficiency, while bash-only enables capable models to combine multiple code modifications in a single tool call. Harness design is thus a conditional systems problem in which each component should be selected for the target model, task type, and resource budget rather than adopted as a default.

\section*{Limitations}

First, our results estimate the conditional effects of the specific harness components and implementations studied here rather than identifying a universally optimal harness. Planning is instantiated through one prompt and update mechanism, and context management follows one threshold-based policy with fixed compaction settings. The action-space intervention is a bundled interface change that jointly varies tool availability, interface-specific prompts, file-state tracking, and automatic post-edit diagnostics, so it does not isolate the effect of tool count or action granularity from the other properties of the interface.
Second, the coverage of the design space and the statistical power are limited. Planning and the action space are ablated only under the default T4/128k configuration due to limited computing resources, and a full factorial study would be required to determine whether their effects persist under other combinations of components and context budgets. Each setting is run once per task, and Terminal-Bench contains only 89 tasks, so many Terminal-Bench contrasts do not reach significance under the paired McNemar test; our conclusions there rest on consistent directions across models and budgets rather than on individually significant cells.
Last, the external validity of our conclusions is bounded by the models and tasks evaluated. We study three sizes from the Nemotron-3 family and Mistral-Medium-3.5-128B on two long-horizon coding benchmarks, of which SWE-Bench Verified is Python-only. Model size is only an imperfect proxy for capability, since differences in training, prior exposure to tool interfaces, and native shell proficiency may also contribute to the observed trends, as Mistral's benchmark-dependent interface preference shows. The crossover points reported here should therefore be validated before being transferred to other model families, harness implementations, or task types outside software engineering.


\bibliographystyle{plainnat}
\bibliography{main}

\clearpage
\beginappendix

\section{Harness Prompts}
\label{app:prompts}

This section reproduces the fixed prompt blocks used during evaluation and explains how they are assembled under different harness configurations. Each model invocation combines an action-interface-specific base prompt with instructions and reminders supplied by the enabled planning and context-management components. The figures reproduce the fixed text, while runtime-dependent fields, including the workspace path, session information, current plan, elision statistics, stored event identifiers, and running summary, are instantiated separately.

\subsection{System prompts}
\label{app:prompts_system}

The action-space intervention changes both the active tool registry and the interface-specific system instructions. Figure~\ref{prompt:base_system} defines the repository-level workflow used with the predefined tool set and explicitly directs the model toward the corresponding exploration, editing, and verification tools. Figure~\ref{prompt:base_system_bash_only} preserves the same high-level workflow and safety requirements while replacing references to unavailable predefined tools with interface-independent instructions. Figure~\ref{prompt:bash_only_system} additionally states that every general workspace operation must be expressed through \texttt{bash}, while auxiliary planning and context-management actions remain controlled by their respective components. Figures~\ref{prompt:tools_system} and~\ref{prompt:tools_system_bash_only} provide the corresponding procedural rules for the tools exposed in each condition. Consequently, the action-space ablation changes both tool availability and the instructions through which those tools are presented to the model.

\begin{figure}[ht]
\small
\centering
\begin{tcolorbox}[
    colback=background,
    colframe=frame
]
\begin{Verbatim}[breaklines=true,breaksymbol=,]
You are a code agent operating inside a local repository.
Your job is to complete the user's coding task with high reliability.

Core workflow:
1. Explore the workspace before acting (list_files, glob_files, grep_text, read_file).
2. Before modifying an existing file you MUST read it first with read_file.
3. Prefer edit_file (targeted old_text/new_text replacement) over write_file (overwrite).
4. Run focused verification (tests, type-check) when appropriate via bash.
5. When the task is done, write a short summary describing what changed and how it was verified.

Rules:
- Never fabricate file contents or test results. Only claim a test passed if you ran it.
- Never operate outside the workspace; paths must be relative to the workspace root.
- For destructive shell commands you will be asked for permission. Don't expect approval.
- Tool calls can be issued in parallel for read-only operations; the harness will batch them.
\end{Verbatim}
\end{tcolorbox}
\caption{Base system prompt used in every configuration with the full action space.}
\label{prompt:base_system}
\end{figure}

\begin{figure}[t]
\small
\centering
\begin{tcolorbox}[
    colback=background,
    colframe=frame
]
\begin{Verbatim}[breaklines=true,breaksymbol=,]
You are a code agent operating inside a local repository.
Your job is to complete the user's coding task with high reliability.

Core workflow:
1. Explore the workspace before acting.
2. Before modifying an existing file you MUST read it first.
3. Prefer surgical edits over rewriting a whole file.
4. Run focused verification (tests, type-check) when appropriate.
5. When the task is done, write a short summary describing what changed and how it was verified.

Rules:
- Never fabricate file contents or test results. Only claim a test passed if you ran it.
- Never operate outside the workspace; paths must be relative to the workspace root.
- For destructive shell commands you will be asked for permission. Don't expect approval.
- Tool calls can be issued in parallel for read-only operations; the harness will batch them.
\end{Verbatim}
\end{tcolorbox}
\caption{Base system prompt used in the bash-only action-space condition. This prompt replaces Figure~\ref{prompt:base_system} at the bash-only configuration.}
\label{prompt:base_system_bash_only}
\end{figure}

\begin{figure}[t]
\small
\centering
\begin{tcolorbox}[
    colback=background,
    colframe=frame
]
\begin{Verbatim}[breaklines=true,breaksymbol=,]
# Working with only a shell

Every interaction with the codebase goes through `bash` — you write the commands yourself.
\end{Verbatim}
\end{tcolorbox}
\caption{Additional system block appended in the bash-only configuration.}
\label{prompt:bash_only_system}
\end{figure}

\begin{figure}[ht]
\small
\centering
\begin{tcolorbox}[
    colback=background,
    colframe=frame  
]
\begin{Verbatim}[breaklines=true,breaksymbol=,]
# Tool Usage Rules

The available tools and their input/output schemas are provided to you via the API's `tools` field. The rules below tell you HOW to use them well.

- Read-only tools (list_files, glob_files, grep_text, read_file, web_fetch) can be called in PARALLEL — issue multiple in a single response when exploring.
- edit_file: small targeted changes. `old_text` must be unique in the file (or pass `replace_all=true`). You MUST read the target with read_file first — the harness rejects edits that skip this.
- write_file: prefer for NEW files. To overwrite an existing file you must pass `overwrite=true` AND have read it fully first.
- bash: tests, lint, git status/diff. Avoid destructive commands — they will be denied by the permission layer.
- web_fetch: for documentation lookup. Don't fetch the same URL repeatedly.

After each tool call, read the result. If it errored, FIX THE INPUT rather than retrying the same call.
\end{Verbatim}
\end{tcolorbox}
\caption{Tool-use rules supplied with the full action space. The block supplements the typed API schemas with instructions for parallel read operations, file modification, shell execution, web access, and recovery from tool errors.}
\label{prompt:tools_system}
\end{figure}

\begin{figure}[t]
\small
\centering
\begin{tcolorbox}[
    colback=background,
    colframe=frame
]
\begin{Verbatim}[breaklines=true,breaksymbol=,]
# Tool Usage Rules

The available tools and their input/output schemas are provided to you via the API's `tools` field. The rules below tell you HOW to use them well.

- bash: tests, lint, git status/diff. Avoid destructive commands — they will be denied by the permission layer.
- update_plan: your live todo list. Re-send the COMPLETE list on every call (it replaces the previous one); keep exactly ONE task `in_progress`.

After each tool call, read the result. If it errored, FIX THE INPUT rather than retrying the same call.
\end{Verbatim}
\end{tcolorbox}
\caption{Tool-use rules supplied in the bash-only action-space condition. The harness generates this block from the active tool registry that also populates the \texttt{tools} field. In the evaluated action-space ablation, the rules for \texttt{bash} and \texttt{update\_plan} remain, while the predefined tools are absent.}
\label{prompt:tools_system_bash_only}
\end{figure}

\subsection{Planning prompts}
\label{app:prompts_planning}

Planning-on is implemented as a persistent scaffold composed of three prompt blocks and an external plan state maintained through \texttt{update\_plan}. 
Figure~\ref{prompt:planning_system} defines the plan-maintenance protocol at the system level and requires an explicit plan for nontrivial tasks.
Figure~\ref{prompt:planning_reminder_first} repeats the initialization requirement only while no plan has been created, encouraging the model to establish the plan before taking another action.
Figure~\ref{prompt:planning_reminder_subsequent} serializes the current stored plan into \texttt{\{PLAN\}} and supplies it anew before each subsequent model invocation rather than appending previous copies to the persistent trajectory. Planning-off removes all three blocks together with the \texttt{update\_plan} tool.

\begin{figure}[ht]
\small
\centering
\begin{tcolorbox}[
    colback=background,
    colframe=frame  
]
\begin{Verbatim}[breaklines=true,breaksymbol=,]
# Task planning

You have an `update_plan` tool that holds a todo list. The harness shows the current plan back to you before every turn, so use it to stay on track.

- For any non-trivial task (about 3+ steps), your FIRST action MUST be to call `update_plan` to break the task into concrete steps.
- Keep it current: mark exactly ONE task `in_progress` while you work on it, and mark a task `completed` the moment it is actually done (don't batch completions).
- Add new tasks as you discover them; re-send the COMPLETE list each time.
- Skip planning only for a single trivial step or a purely informational request.
\end{Verbatim}
\end{tcolorbox}
\caption{Planning system prompt. It requires the model to create an explicit task plan for nontrivial requests, maintain exactly one active step, and update task status as execution progresses.}
\label{prompt:planning_system}
\end{figure}

\begin{figure}[ht]
\small
\centering
\begin{tcolorbox}[
    colback=background,
    colframe=frame  
]
\begin{Verbatim}[breaklines=true,breaksymbol=,]
<system-reminder>
You have not created a plan yet. Before taking any other action, call update_plan to break this task into concrete steps. (Skip planning only for a single trivial step or a purely informational request.)
</system-reminder>
\end{Verbatim}
\end{tcolorbox}
\caption{Planning reminder inserted on the first turn when no plan has yet been created. It requests a call to \texttt{update\_plan} before any other action unless the task
is trivial or purely informational.}
\label{prompt:planning_reminder_first}
\end{figure}

\begin{figure}[ht]
\small
\centering
\begin{tcolorbox}[
    colback=background, 
    colframe=frame  
]
\begin{Verbatim}[breaklines=true,breaksymbol=,]
<system-reminder>
Current plan (update it via update_plan as you progress; keep exactly one task in_progress, and mark a task completed the moment it is actually done):
{PLAN}
</system-reminder>
\end{Verbatim}
\end{tcolorbox}
\caption{Planning reminder inserted before each subsequent model turn. The placeholder is replaced by the current stored plan, which is supplied anew rather than appended to the persistent context history.}
\label{prompt:planning_reminder_subsequent}
\end{figure}

\subsection{Context-management prompts}
\label{app:prompts_context}

The context-management templates operate at two stages of the harness. Figure~\ref{prompt:summarizer} shows the prompt issued in a separate LLM call when M3 summarization is triggered. The harness supplies the oldest selected middle-region events together with any existing running summary, and the resulting text replaces those events in subsequent agent inputs. Figure~\ref{prompt:context_fragments} shows how elided observations and the running summary are represented to the task-solving model. T1 receives an unrecoverable elision stub, whereas T2 and T4 additionally receive the stored event identifier required by \texttt{recall\_event}. The running summary is inserted immediately after the preserved preamble.

\begin{figure}[t]
\small
\centering
\begin{tcolorbox}[
    colback=background,
    colframe=frame
]
\begin{Verbatim}[breaklines=true,breaksymbol=,]
You are compacting a coding agent's conversation to save context. You are given the OLDEST part of the transcript (and possibly a previous summary to update).

Respond with TEXT ONLY — do NOT call any tools. Produce a concise structured summary under these exact headings (omit a heading only if truly empty):

## Goal
The user's task / intent (preserve it precisely).
## Files touched
Paths examined or modified, with the key changes made and why.
## Done
What has been accomplished, with concrete results (tests passing, bug located, …).
## Pending
What still needs doing.
## Errors & fixes
Errors hit and how they were resolved (or are still open).
## Current state
Key variables / branch / test status / anything needed to continue.
## Next step
The immediate next action, in line with the most recent work.

Be specific (keep exact file paths, function names, error messages, task ids). If a previous summary is given, UPDATE it: keep still-true facts, drop stale ones, merge in the new events. The raw events remain retrievable, so summarize — don't transcribe.
\end{Verbatim}
\end{tcolorbox}
\caption{Prompt used by the summarization mechanism (M3). The harness invokes the same model in a separate call over the oldest selected events in the middle region. The fixed headings support incremental maintenance of the running summary because the previous summary is supplied to the model and revised with the newly selected events.}
\label{prompt:summarizer}
\end{figure}

\begin{figure}[t]
\small
\centering
\begin{tcolorbox}[
    colback=background,
    colframe=frame
]
\begin{Verbatim}[breaklines=true,breaksymbol=,]
### Stub left in place of an elided tool observation (M1 with recall, i.e. T2 and T4)

[tool output elided: {N_LINES} lines / {N_CHARS} chars. Use recall_event({EVENT_ID}) for the full output, or re-read/re-run.]

### Stub left in place of an elided tool observation (M1 without recall, i.e. T1)

[tool output elided: {N_LINES} lines / {N_CHARS} chars. Re-read or re-run to get it again.]

### Wrapper around the running summary (M3, i.e. T3 and T4)

<context-summary>
{SUMMARY}
(Older events were summarized. Their raw content is recallable via recall_event up to event {EVENT_ID}.)
</context-summary>
\end{Verbatim}
\end{tcolorbox}
\caption{Text fragments inserted by context management. The first two fragments replace an elided observation and report the amount of removed content. The variant used when M2 is enabled also provides the event identifier for \texttt{recall\_event}. The third fragment wraps the running summary, which replaces the events it covers and is inserted immediately after the preamble.}
\label{prompt:context_fragments}
\end{figure}

\subsection{Stuck-detection prompts}
\label{app:prompts_stuck}


Stuck detection is enabled identically in every experimental condition and therefore belongs to the fixed execution substrate. The harness defines a streak as the run of most recent tool calls that share the same tool name and byte-identical arguments; a call with different arguments, such as a paginated read at a new offset, ends the streak. Figure~\ref{prompt:stuck_reminder} shows the two reminders, selected according to whether the repeated calls merely return an existing result or keep failing in the same way. A reminder is inserted once per streak when the streak reaches five identical calls of any status, or five identical failing calls; at runtime, \texttt{\{TOOL\}} and \texttt{\{N\}} are replaced by the repeated tool name and the current streak length. If an identical failing streak continues to eight calls, the run is terminated with a dedicated stop reason rather than running to the step budget. Permission denials do not count as failures for this purpose.

\begin{figure}[t]
\small
\centering
\begin{tcolorbox}[
    colback=background,
    colframe=frame
]
\begin{Verbatim}[breaklines=true,breaksymbol=,]
### Injected when the same call repeats and keeps failing

<system-reminder>
You have called `{TOOL}` with the SAME arguments {N} times and it keeps failing the same way. Repeating it will NOT work. STOP — read the actual error, then try a DIFFERENT command, inspect more context, or step back and reconsider your plan. Do NOT issue the same call again.
</system-reminder>

### Injected when the same call merely repeats

<system-reminder>
You have called `{TOOL}` with the SAME arguments {N} times. You're not making progress — you already have this result. Move on to the next concrete step instead of repeating it.
</system-reminder>
\end{Verbatim}
\end{tcolorbox}
\caption{Reminders used by the stuck-detection mechanism. The harness identifies streaks of calls with the same tool name and arguments and selects the reminder according to whether the calls return the same failure. A reminder is inserted once per streak. Continued growth of an identical failing streak terminates the run before it exhausts the step budget.}
\label{prompt:stuck_reminder}
\end{figure}

\section{Tool Descriptions}
\label{app:tools}

Table~\ref{tab:tools} summarizes tool availability and principal arguments. Each model-facing tool definition consists of a tool name, a typed argument schema, and a natural-language description. The figures below reproduce the exact description strings used during evaluation, including their usage protocols, error behavior, and side effects. The predefined-tool condition exposes the file, search, execution, and web interfaces, whereas the bash-only condition removes the predefined workspace tools and retains \texttt{bash}. The auxiliary tools \texttt{update\_plan} and \texttt{recall\_event} are controlled independently by the planning and context-management settings.

\subsection{File input and output}

Figures~\ref{prompt:tool_read_file}--\ref{prompt:tool_edit_file} reproduce the three structured file operations available in the predefined-tool condition. Figure~\ref{prompt:tool_read_file} defines bounded, line-numbered file access and records a complete read in the session-level file state. Figure~\ref{prompt:tool_write_file} creates new files or performs explicitly authorized full-file replacement, while Figure~\ref{prompt:tool_edit_file} applies byte-exact targeted replacements to existing files. Both mutation tools require an eligible recorded read before modifying an existing file and update the harness file state after success. Equivalent modifications issued through \texttt{bash} do not participate in this state-tracking and automatic-diagnostic path.

\begin{figure}[t]
\small
\centering
\begin{tcolorbox}[
    colback=background,
    colframe=frame
]
\begin{Verbatim}[breaklines=true,breaksymbol=,]
Read the contents of a text file from the workspace, returned with 1-based line numbers and a header describing what range was read. 
When: call this BEFORE edit_file or write_file(overwrite=True) on any existing file — the harness enforces read-before-write and rejects edits without a recorded full read. For directory listings use list_files; for content search use grep_text. 
Protocol: returns lines [offset, offset+limit) using 1-based offsets. Default reads from line 1 with limit=2000. The read is recorded as `full_read=True` ONLY when offset=1 AND limit covers the entire file; partial reads are recorded as `full_read=False` and do NOT satisfy the read-before-write gate for edit_file/write_file(overwrite=True). Encoding is UTF-8 with the replacement character for invalid bytes. 
Error modes: returns an error for non-existent paths, paths outside the workspace, directories (suggests list_files), binary files (NUL byte in first 8KB), files exceeding the configured `max_read_bytes` cap (suggests `bash` with head/sed instead), or offsets past EOF. Sensitive paths (`.env`, secrets) trigger a permission prompt via the configured approval callback. 
Side effects: records the read in the session's file state. Subsequent edit_file / write_file calls on this path in this session satisfy the read-before-write check IF full_read=True. The recorded state survives across turns of the same session.
\end{Verbatim}
\end{tcolorbox}
\caption{Tool-call description for \texttt{read\_file}.}
\label{prompt:tool_read_file}
\end{figure}

\begin{figure}[t]
\small
\centering
\begin{tcolorbox}[
    colback=background,
    colframe=frame
]
\begin{Verbatim}[breaklines=true,breaksymbol=,]
Write content to a file path — creates a new file or overwrites an existing one. Returns a unified diff of the change (against empty content for new files).
When: use this for (a) creating new files, (b) full-file rewrites of existing files. For small in-place edits to existing files use edit_file instead. New-file creation does not require overwrite=true; only overwriting an existing file does.
Protocol: if the path doesn't exist, the file is created and parent directories are created as needed. If the path exists, you MUST pass overwrite=true AND you MUST have called read_file with full_read=True on this path in this session first — overwriting an unread or partial-read file is rejected to prevent silent loss of user changes. The diff is shown to the user via the permission callback before being applied. 
Error modes: returns a recoverable error when the file exists and overwrite=false (suggests edit_file or overwrite=true), or when the path is a directory. RAISES (caught by harness) for: path outside the workspace, existing file never read or only partial-read, file modified externally since the recorded read, or user denying the write via the permission callback. 
Side effects: writes the file to disk and creates parent directories as needed. Updates the session's file state with the new content (full_read=True), so a subsequent edit_file on this path in the same session passes the read-before-write check without an explicit read. The path is recorded as a `changed_file` in the agent result.
\end{Verbatim}
\end{tcolorbox}
\caption{Tool-call description for \texttt{write\_file}.}
\label{prompt:tool_write_file}
\end{figure}

\begin{figure}[t]
\small
\centering
\begin{tcolorbox}[
    colback=background,
    colframe=frame
]
\begin{Verbatim}[breaklines=true,breaksymbol=,]
Edit a text file by replacing one or more occurrences of `old_text` with `new_text`. Returns a unified diff of the applied change. 
When: use this for SMALL, TARGETED changes to EXISTING files. For new files use write_file; for full-file rewrites use write_file(overwrite=true). 
Protocol: `old_text` MUST appear in the file at least once and the match is byte-exact (including whitespace and line endings). If `old_text` appears more than once, either provide MORE surrounding context to make it unique, OR pass `replace_all=true`. `old_text == new_text` is rejected as a no-op. The file MUST have been read with full_read=True in this session BEFORE calling edit_file — the harness enforces this and rejects edits without a recorded full read. The diff is shown to the user via the permission callback before being applied. 
Error modes: returns a recoverable error for: non-existent path (suggests write_file), path is a directory, `old_text` not found, ambiguous match without `replace_all`, identical `old_text`/`new_text`, or non-UTF-8 content. RAISES (caught by harness, surfaced to the model as a ToolMessage error) for: path outside the workspace, file never read or only partially read, file modified externally since the recorded read, or user denying the edit via the permission callback.
Side effects: writes the modified content to disk. Updates the session's file state with the new content (full_read=True), so a subsequent edit_file or write_file(overwrite=true) on this path in the same session passes the read-before-write check without re-reading. The path is recorded as a `changed_file` in the agent result for this turn.
\end{Verbatim}
\end{tcolorbox}
\caption{Tool-call description for \texttt{edit\_file}.}
\label{prompt:tool_edit_file}
\end{figure}

\subsection{Search}

Figures~\ref{prompt:tool_list_files}--\ref{prompt:tool_grep_text} reproduce the three read-only discovery interfaces available in the predefined-tool condition. Figure~\ref{prompt:tool_list_files} returns a bounded tree representation of a directory, Figure~\ref{prompt:tool_glob_files} locates files by path pattern, and Figure~\ref{prompt:tool_grep_text} searches file contents by regular expression. These tools do not mutate the workspace or file state, may be invoked concurrently when independent, and consistently apply the configured workspace-ignore patterns.

\begin{figure}[t]
\small
\centering
\begin{tcolorbox}[
    colback=background,
    colframe=frame
]
\begin{Verbatim}[breaklines=true,breaksymbol=,]
List files and directories in the workspace, returning a tree-style view with relative paths and (for files) byte sizes. 
When: use this to explore project structure before reading specific files. For pattern-based file discovery use glob_files; for content search use grep_text. 
Protocol: returns up to `max_entries` entries (default 200, max 2000), sorted alphabetically within each directory. Set `recursive=true` to descend into subdirectories — depth is reflected in output indentation. 
Error modes: returns an error if `path` doesn't exist or isn't a directory. Workspace ignore patterns (`.git/**`, `__pycache__/**`, `.venv/**`, `node_modules/**`) are silently filtered from output even with recursive=true. Paths outside the workspace are rejected. 
Side effects: none — read-only, does not affect file_state or trigger any permission caching.
\end{Verbatim}
\end{tcolorbox}
\caption{Tool-call description for \texttt{list\_files}.}
\label{prompt:tool_list_files}
\end{figure}

\begin{figure}[t]
\small
\centering
\begin{tcolorbox}[
    colback=background,
    colframe=frame
]
\begin{Verbatim}[breaklines=true,breaksymbol=,]
Find files matching a glob pattern within the workspace. 
When: use this when you know the file pattern (`**/*.py`, `src/components/*.tsx`). For listing all entries in a single directory use list_files; for content search use grep_text.
Protocol: returns up to `max_matches` matching file paths (default 200, max 1000), workspace-relative. Pattern syntax: `**` recursive, `*` within segment, `?` single char, `[abc]` character class. Only files are returned — directories matching the pattern are excluded. Error modes: returns an empty result (not an error) when no files match. Returns an error if `path` doesn't exist or isn't a directory. Workspace ignore patterns (`.git/**`, etc.) are filtered silently.
Side effects: none — read-only, does not affect file_state.
\end{Verbatim}
\end{tcolorbox}
\caption{Tool-call description for \texttt{glob\_files}.}
\label{prompt:tool_glob_files}
\end{figure}

\begin{figure}[t]
\small
\centering
\begin{tcolorbox}[
    colback=background,
    colframe=frame
]
\begin{Verbatim}[breaklines=true,breaksymbol=,]
Search for a regex pattern across workspace files, returning matched lines with paths and line numbers. 
When: use this when you need to find code by content — function definitions, error messages, configuration values, TODOs. For pattern-based file discovery (no content matching) use glob_files. 
Protocol: `query` is a regex (uses ripgrep when available, else Python `re`; both support standard regex features). Returns up to `max_matches` hits (default 100), formatted as `path:line:matched_text`. Use `include` to restrict to a filename glob (e.g., `*.py`). Set `case_sensitive=false` for case-insensitive matching (implemented via `(?i)` inline flag — portable across backends). 
Error modes: returns empty result (not error) when no matches. Binary files (NUL byte in first 8KB) and files >10MB are silently skipped on both backend paths. Workspace ignore patterns are respected. Side effects: none — read-only, does not affect file_state.
\end{Verbatim}
\end{tcolorbox}
\caption{Tool-call description for \texttt{grep\_text}.}
\label{prompt:tool_grep_text}
\end{figure}

\subsection{Execution and web access}

Figures~\ref{prompt:tool_bash} and~\ref{prompt:tool_bash_only} show the alternative model-facing descriptions attached to the same shell executor in the two action-space conditions. In both conditions, commands are executed through \texttt{/bin/sh -c}. The predefined-tool description directs common file and search operations toward the structured interfaces, whereas the bash-only description requires all general workspace interaction to be expressed through shell commands and asks the model to validate modifications explicitly. Figure~\ref{prompt:tool_web_fetch} describes concrete-URL retrieval, which is available only in the predefined-tool condition.

\begin{figure}[t]
\small
\centering
\begin{tcolorbox}[
    colback=background,
    colframe=frame
]
\begin{Verbatim}[breaklines=true,breaksymbol=,]
Execute a shell command in the workspace and return its exit code, stdout, and stderr. Useful for running tests, linters, build commands, and read-only git queries. 
When: use this when you need to invoke external programs — pytest, ruff, mypy, npm test, git diff, etc. For file reading prefer read_file (it records read state for read-before-write); for file search prefer grep_text.
Protocol: command is run via `/bin/sh -c` so pipes, redirects, and shell builtins work. cwd defaults to the workspace root and must be inside the workspace. `timeout_seconds` (default 120, max 600) kills the process on expiry and returns whatever stdout/stderr it produced before the kill, with timed_out=True. The exit code is returned literally; non-zero exit codes produce an `is_error=True` ToolMessage so the model sees the failure clearly. v1 does NOT track which files the shell touched — run `git status` or `git diff` afterwards if you need to know. 
Error modes: the PermissionManager rejects destructive commands by default — `rm -rf`, `sudo`, `git push`, `git reset --hard`, `chmod -R`, `chown -R` all return a denied error. Don't expect them to be approved. Read-only commands in the default allow list — `git status`, `git diff`, `git log`, `git branch`, `git show`, `ls`, `pwd`, `rg`, `cat`, `head`, `tail`, `wc`, `file` — run without prompting. Everything else goes through the approval callback. Returns a recoverable error for non-existent or non-directory cwd. RAISES (harness-caught) for cwd outside the workspace or user denying the command via the callback. 
Side effects: anything the command does — file creation, network calls, package installs, git operations. The agent harness does NOT track these as `changed_files` (use `git status` after to discover modifications). The session's file_state is NOT updated, so files modified by bash are NOT automatically eligible for edit_file without a fresh read_file.
\end{Verbatim}
\end{tcolorbox}
\caption{Tool-call description supplied for \texttt{bash} in the predefined
tool-set condition.}
\label{prompt:tool_bash}
\end{figure}

\begin{figure}[t]
\small
\centering
\begin{tcolorbox}[
    colback=background,
    colframe=frame
]
\begin{Verbatim}[breaklines=true,breaksymbol=,]
Execute a shell command in the workspace and return its exit code, stdout, and stderr. Useful for running tests, linters, build commands, and read-only git queries. This is your ONLY tool for touching the codebase — reading, searching and editing files all happen through commands you write.
Protocol: command is run via `/bin/sh -c` so pipes, redirects, and shell builtins work. cwd defaults to the workspace root and must be inside the workspace. `timeout_seconds` (default 120, max 600) kills the process on expiry and returns whatever stdout/stderr it produced before the kill, with timed_out=True. The exit code is returned literally; non-zero exit codes produce an `is_error=True` ToolMessage so the model sees the failure clearly. v1 does NOT track which files the shell touched — run `git status` or `git diff` afterwards if you need to know. 
Error modes: the PermissionManager rejects destructive commands by default — `rm -rf`, `sudo`, `git push`, `git reset --hard`, `chmod -R`, `chown -R` all return a denied error. Don't expect them to be approved. Read-only commands in the default allow list — `git status`, `git diff`, `git log`, `git branch`, `git show`, `ls`, `pwd`, `rg`, `cat`, `head`, `tail`, `wc`, `file` — run without prompting. Everything else goes through the approval callback. Returns a recoverable error for non-existent or non-directory cwd. RAISES (harness-caught) for cwd outside the workspace or user denying the command via the callback. Nothing validates your edits, so after modifying a file read it back to confirm the change landed as intended.
\end{Verbatim}
\end{tcolorbox}
\caption{Replacement model-facing description of \texttt{bash} in the bash-only condition.}
\label{prompt:tool_bash_only}
\end{figure}

\begin{figure}[t]
\small
\centering
\begin{tcolorbox}[
    colback=background,
    colframe=frame
]
\begin{Verbatim}[breaklines=true,breaksymbol=,]
Fetch the content of an HTTP(S) URL and return it as readable text. HTML responses are converted to clean markdown by default; JSON and text responses pass through unchanged. 
When: use this to read documentation pages, API responses, GitHub READMEs, library docs, blog posts — anything you have a concrete URL for. To DISCOVER URLs by topic use web_search first. For files inside the workspace use read_file. 
Protocol: URL must be `http://` or `https://` (other schemes rejected at schema validation). Response is capped at `max_bytes` (default 2MB, max 10MB); larger responses are truncated and flagged in metadata. `format`: `markdown` (default — HTML->markdown via markdownify, scripts and styles stripped), `text` (HTML->plain text via BeautifulSoup), `html` (raw). Non-HTML responses ignore `format` and are decoded as UTF-8 with the replacement character for invalid bytes. Redirects are followed automatically; `final_url` in the response header shows where you actually ended up. 
Error modes: HTTP 4xx/5xx responses return is_error=True with a 500-char body preview so the model can decide whether to retry / pick a different URL. Network failures (DNS, refused, TLS, timeout) also return is_error=True (network is unreliable — recoverable error, not raise). Binary content (images, archives, ...) returns a notice with size and content-type rather than garbled bytes — use bash with curl if you genuinely need byte access. RAISES (harness-caught) only for user denying the URL via the permission callback. 
Side effects: makes one outbound HTTP request to the URL. No workspace files are touched. No file_state is recorded. The agent result does not list any changed_files.
\end{Verbatim}
\end{tcolorbox}
\caption{Tool-call description for \texttt{web\_fetch}.}
\label{prompt:tool_web_fetch}
\end{figure}

\subsection{Planning and context recall}

The tools in this subsection are auxiliary component interfaces rather than predefined workspace actions. Figure~\ref{prompt:tool_update_plan} describes \texttt{update\_plan}, which is exposed whenever planning is enabled and replaces the complete in-memory plan without modifying workspace files. Figure~\ref{prompt:tool_recall_event} describes \texttt{recall\_event}, which is exposed only under T2 and T4 and retrieves stored historical content by event identifier. When enabled, both tools remain available under either action-space condition.

\begin{figure}[t]
\small
\centering
\begin{tcolorbox}[
    colback=background,
    colframe=frame
]
\begin{Verbatim}[breaklines=true,breaksymbol=,]
Create or update your task plan (a todo list). The harness shows this plan back to you before every turn so you stay on track and don't lose steps. 
When: for any non-trivial task (about 3+ steps), call this FIRST to lay out the steps, then keep it updated as you work. Skip only for a single trivial step or a purely informational request. 
Protocol: pass the COMPLETE updated list every time — it fully replaces the previous list (no incremental edits). Each item has `content` (imperative, e.g. 'Fix the parser in foo.py'), `status` (pending | in_progress | completed), and `activeForm` (present-continuous, e.g. 'Fixing the parser'). Keep EXACTLY ONE task in_progress at a time; mark a task completed the moment it is actually done (do not batch completions); add new tasks as you discover them. 
Side effects: updates the in-memory plan only — it touches no files and is not written to disk.
\end{Verbatim}
\end{tcolorbox}
\caption{Tool-call description for \texttt{update\_plan}.}
\label{prompt:tool_update_plan}
\end{figure}

\begin{figure}[t]
\small
\centering
\begin{tcolorbox}[
    colback=background,
    colframe=frame
]
\begin{Verbatim}[breaklines=true,breaksymbol=,]
Fetch the full original content of a past turn (event) by its id. When the conversation has been compacted, old tool outputs are replaced by stubs like '[tool output elided ... Use recall_event(41) ...]' and the oldest turns may be folded into a summary; call this with the event id to get the verbatim messages (including the full tool output) back. Often you can instead just re-read the file or re-run the command — use recall_event when the output isn't easily reproducible (e.g. a past test log). Read-only; returns an error for an out-of-range id.
\end{Verbatim}
\end{tcolorbox}
\caption{Tool-call description for \texttt{recall\_event}.}
\label{prompt:tool_recall_event}
\end{figure}

\section{Trajectory Analysis}
\label{app:traj_analysis}

Rather than evaluating harnesses solely by final task success, we decompose each agent trajectory into interpretable behavioral and failure categories. These analyses examine trajectory survival, behavioral composition, action granularity, and the stage at which unresolved runs terminate. We use them to assess whether the observed behavioral changes are consistent with the mechanisms proposed in \S\ref{sec:analysis}. The appendix is organized around three complementary annotation schemes, followed by judge validation, benchmark-specific trajectory results, summary statistics, recall-event usage, and the full judge prompts.

\subsection{Annotation Schemes}
We use three complementary annotation schemes. Failure-stage attribution identifies the earliest failed stage in unresolved SWE-Bench repair trajectories. The two behavior encodings describe how agents allocate actions across the problem-solving process in SWE-Bench and Terminal Bench.

\subsubsection{SWE-Bench Failure-Stage Attribution}
For unresolved SWE-Bench trajectories, we assign a failure stage corresponding to the earliest point at which the repair process fails. The stages follow the natural debugging pipeline: file localization, line localization, patch implementation, and verification. Using the earliest-failure convention ensures that downstream symptoms, such as failed tests, do not obscure earlier localization or implementation errors.

\subsubsection{SWE-Bench Behavior Encoding}
We use SWE-Bench behavior encoding to characterize the agent's primary workflow purpose at each turn. Building on the trajectory-analysis framework of \citet{mehtiyev2026beyond}, we assign every turn to one of five phases: localization, reproduction, fixing, verification, and other auxiliary behavior. These labels capture how the agent distributes effort across the debugging process; the five phase symbols and their trajectory sources are summarized in Table~\ref{tab:traj_encoding}.
\begin{table}[h]
\centering
\small
\begin{tabular}{@{}llll@{}}
\toprule
\textbf{Category} & \textbf{Sym.} & \textbf{Meaning} & \textbf{Source in trajectory} \\
\midrule
Localize  & L & Locate relevant code   & file read, \texttt{grep}, \texttt{glob}, \texttt{list}, \texttt{find} \\
Reproduce & R & Build/run repro script & create / edit / run a \texttt{reproduce} file \\
Fix       & F & Patch source           & \texttt{edit} on a repository file \\
Verify    & V & Run validation         & test run (pass, fail, or error) \\
Other     & O & Setup and auxiliary    & \texttt{pip}, \texttt{conda}, \texttt{recall\_event}, \texttt{submit}, \ldots \\
\bottomrule
\end{tabular}
\caption{SWE-Bench trajectory encoding: five phase symbols.}
\label{tab:traj_encoding}
\end{table}

\subsubsection{Terminal-Bench Action-Level Encoding}
We use Terminal-Bench action-level encoding to classify the semantic intent of each shell-based action. Unlike the SWE-Bench encoding, which labels whole turns, this scheme labels each tool call individually. We define 10 fine-grained action-type symbols and group them into four higher-level phases: understanding, code writing, verification, and other auxiliary operations. The complete 10-symbol action taxonomy is shown in Table~\ref{tab:tbench_purpose}.

\begin{table}[h]
\centering
\begin{tabular}{@{}llll@{}}
\toprule
\textbf{Category} & \textbf{Symbol} & \textbf{Meaning} & \textbf{Source in trajectory} \\
\midrule
\multirow{2}{*}{Understand}
 & I & Inspect file content    & \texttt{cat}, \texttt{head}, \texttt{tail}, \texttt{less}, \texttt{sed -n} \\
 & S & Search / discover       & \texttt{ls}, \texttt{find}, \texttt{grep}\,|\,\texttt{rg}, \texttt{which} \\
\midrule
\multirow{3}{*}{Write code}
 & C & Create (new file)       & heredoc write, \texttt{touch}, \texttt{tee} to a new path \\
 & M & Modify (existing file)  & \texttt{sed -i}, \texttt{patch}, append redirect \\
 & E & Environment setup       & \texttt{pip}, \texttt{apt}, \texttt{conda}, \texttt{make}, build \\
\midrule
\multirow{3}{*}{Verify}
 & X & Execute own work         & \texttt{python foo.py}, \texttt{./run.sh}, compiled binary \\
 & T & Test (real framework)    & \texttt{pytest}, \texttt{unittest}, \texttt{npm test}, \texttt{go test} \\
 & V & Verify produced artifact & \texttt{diff}, \texttt{md5sum}, re-read of output file \\
\midrule
\multirow{2}{*}{Other}
 & N & Navigate        & \texttt{cd}, \texttt{pushd}, \texttt{pwd} as the sole call \\
 & G & General / other & \texttt{git}, \texttt{echo}, \texttt{export}, \texttt{rm}, shell plumbing, \ldots \\
\bottomrule
\end{tabular}
\caption{Terminal-Bench trajectory encoding: 10 action-type symbols grouped into four phases. }
\label{tab:tbench_purpose}
\end{table}

\subsection{Judge Setup and Human Validation}
\label{app:human_eval}
We use LLM-based judges for SWE-Bench turn-purpose classification, SWE-Bench failure-stage attribution, and Terminal-Bench action-purpose classification. The corresponding prompts are reproduced in Figures~\ref{prompt:turn_purposes}, \ref{prompt:failure_stage}, and~\ref{prompt:action_purpose}. All judge calls use GPT-5.5 with reasoning effort set to high and a decoding temperature of 0.6.

We validate the LLM-judge annotations with three human annotators. A sample of 200 trajectories is divided evenly into three non-overlapping splits, one per annotator, with one annotator labeling one more trajectory than the other two; each annotator labels every unit in their split. The validation covers 15,610 labeled units in total, including 5,306 Terminal-Bench action labels, 10,254 SWE-Bench action labels, and 50 SWE-Bench failure-stage diagnoses. For each split and unit type, annotators independently assign labels using the same inventories provided to the LLM judge. We compare the human labels with the judge outputs using raw agreement and Cohen's $\kappa$.

Overall judge--human agreement is high across annotators. The per-split overall agreement ranges from 88.7\% to 98.4\%, with Cohen's $\kappa$ ranging from 0.858 to 0.980. Aggregating across the three splits gives approximately 94.2\% raw agreement and a weighted mean Cohen's $\kappa$ of 0.929. Agreement is especially strong for SWE-Bench action labels, where raw agreement ranges from 92.7\% to 99.0\% and $\kappa$ ranges from 0.881 to 0.984. Terminal-Bench action labels also show substantial agreement, with raw agreement ranging from 79.8\% to 97.3\% and $\kappa$ ranging from 0.758 to 0.965. For failure-stage diagnosis, agreement remains high, ranging from 87.5\% to 100.0\%, with $\kappa$ ranging from 0.813 to 1.000.

These results indicate that the LLM judge is closely aligned with independent human annotations across both benchmark-specific action taxonomies and failure-stage labels. They support the use of judge-produced trajectory annotations for scaling the behavioral analysis in \S\ref{sec:analysis} to the full set of experimental trajectories.

\subsection{SWE-Bench Trajectory Results}
\subsubsection{Failure-Stage Results}

The SWE-Bench failure-stage judge is applied only to unresolved trajectories. It receives the issue description, the source files modified by the gold patch, the agent's source-only patch, and the final portion of the trajectory, and then assigns the earliest stage that was not completed correctly. Table~\ref{tab:app_failure_stage} reports the resulting distribution over the four stages for every 128k setting. The complementary termination statistics, which are derived from the behavior encodings rather than from the judge and record whether a run edited a file before terminating and which phase it was in when it stopped, are reported separately in Table~\ref{tab:app_termination_stage_full}.

Three patterns stand out. First, the dominant failure stage shifts with model capability: for Nemotron-3 30B, more than half of unresolved runs fail at file localization, whereas for Nemotron-3 550B and Mistral-Medium-3.5-128B, the majority reach the correct file and lines but fail at patch implementation. Second, removing planning or the predefined tools from Nemotron-3 30B pushes the file-localization share from 56.3\% to 73.8\% and 76.6\%, respectively, consistent with the premature-termination behavior described in \S\ref{sec:analysis}. Third, the Mistral bash-only setting is the only strong-model configuration in which file-localization failures rise sharply, from 16.0\% to 41.4\%, which indicates that its 23.2-point drop under bash-only originates mainly before the repair step. Across the context-management tiers, the stage distribution changes little, so the tier comparison at 128k reflects small differences in how many runs fail rather than where they fail.

\begin{table}[t]
\centering
\caption{Failure-stage distribution of unresolved SWE-Bench runs at a 128k context-window budget, as assigned by the LLM judge (Figure~\ref{prompt:failure_stage}). \emph{Unres.}\ is the number of unresolved runs out of 500. The four stage columns give the percentage of unresolved runs whose earliest failed stage is file localization, line localization, patch implementation, or verification, and sum to 100 within each row; the few unresolved runs whose trajectory could not be scanned by the judge (at most three per setting) are excluded from the percentages.}
\label{tab:app_failure_stage}
\small
\resizebox{\linewidth}{!}{%
\begin{tabular}{@{}llrrrrr@{}}
\toprule
 & & & \multicolumn{4}{c}{Earliest failed stage (\%)} \\
\cmidrule(lr){4-7}
Model & Setting & Unres. & File loc. & Line loc. & Patch & Verif. \\
\midrule
\multirow{7}{*}{Nemotron-3 30B} & T0 & 376 & 51.3 & 13.0 & 31.9 & 3.7 \\
 & T1 & 375 & 49.5 & 12.6 & 35.8 & 2.1 \\
 & T2 & 370 & 54.5 & 12.5 & 30.5 & 2.5 \\
 & T3 & 382 & 55.1 & 9.2 & 32.5 & 3.1 \\
 & T4 & 374 & 56.3 & 9.1 & 32.4 & 2.1 \\
\cmidrule(lr){2-7}
 & T4 w/o plan & 432 & 73.8 & 7.0 & 18.1 & 1.2 \\
 & T4 bash only & 449 & 76.6 & 9.2 & 13.4 & 0.9 \\
\cmidrule(lr){1-7}
\multirow{7}{*}{Nemotron-3 120B} & T0 & 299 & 40.1 & 12.5 & 45.8 & 1.7 \\
 & T1 & 278 & 42.6 & 12.3 & 42.2 & 2.9 \\
 & T2 & 274 & 37.2 & 13.1 & 47.4 & 2.2 \\
 & T3 & 280 & 38.6 & 11.4 & 46.8 & 3.2 \\
 & T4 & 280 & 44.3 & 8.9 & 44.6 & 2.1 \\
\cmidrule(lr){2-7}
 & T4 w/o plan & 267 & 40.1 & 13.5 & 43.1 & 3.4 \\
 & T4 bash only & 288 & 42.9 & 11.1 & 40.1 & 5.9 \\
\cmidrule(lr){1-7}
\multirow{7}{*}{Nemotron-3 550B} & T0 & 201 & 20.9 & 13.4 & 60.2 & 5.5 \\
 & T1 & 174 & 22.4 & 13.2 & 61.5 & 2.9 \\
 & T2 & 163 & 17.2 & 19.6 & 58.9 & 4.3 \\
 & T3 & 171 & 21.1 & 15.8 & 57.9 & 5.3 \\
 & T4 & 171 & 16.4 & 11.7 & 64.9 & 7.0 \\
\cmidrule(lr){2-7}
 & T4 w/o plan & 161 & 17.5 & 16.9 & 57.5 & 8.1 \\
 & T4 bash only & 153 & 20.5 & 14.6 & 57.0 & 7.9 \\
\cmidrule(lr){1-7}
\multirow{7}{*}{Mistral-Medium-3.5-128B} & T0 & 163 & 20.2 & 12.9 & 54.6 & 12.3 \\
 & T1 & 157 & 14.7 & 18.6 & 60.9 & 5.8 \\
 & T2 & 165 & 19.4 & 10.3 & 63.0 & 7.3 \\
 & T3 & 167 & 22.8 & 17.4 & 52.7 & 7.2 \\
 & T4 & 157 & 16.0 & 16.7 & 60.3 & 7.1 \\
\cmidrule(lr){2-7}
 & T4 w/o plan & 155 & 20.3 & 17.6 & 56.2 & 5.9 \\
 & T4 bash only & 273 & 41.4 & 9.2 & 45.4 & 4.0 \\
\bottomrule
\end{tabular}%
}
\end{table}

\subsubsection{Behavior Profiles}

\begin{figure}[h]
\centering
\includegraphics[width=\textwidth]{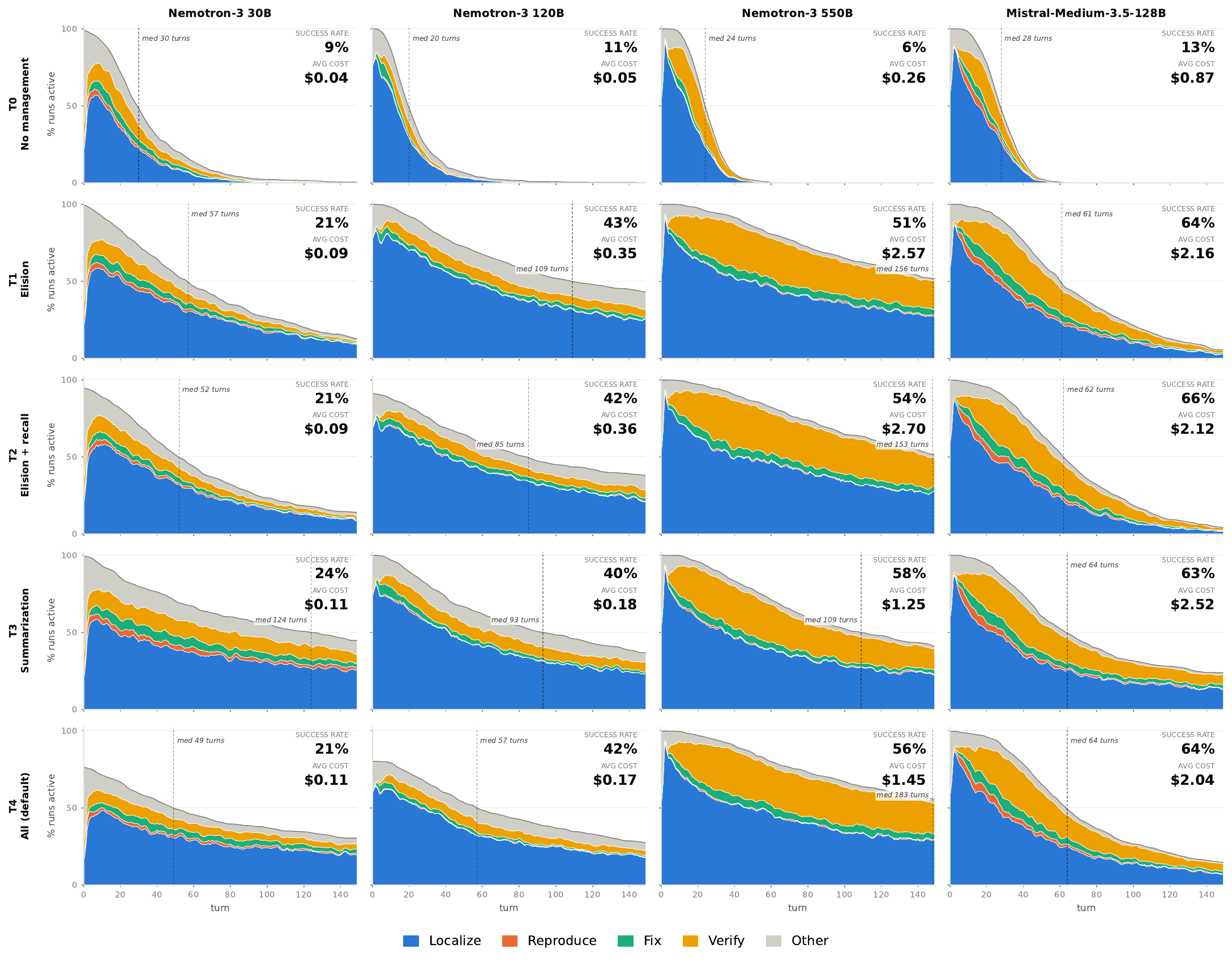}
\caption{SWE-Bench trajectory profiles across context-management strategies at 32k.}
\label{fig:swe_behavior_ctx32k}
\end{figure}

\begin{figure}[h]
\centering
\includegraphics[width=\textwidth]{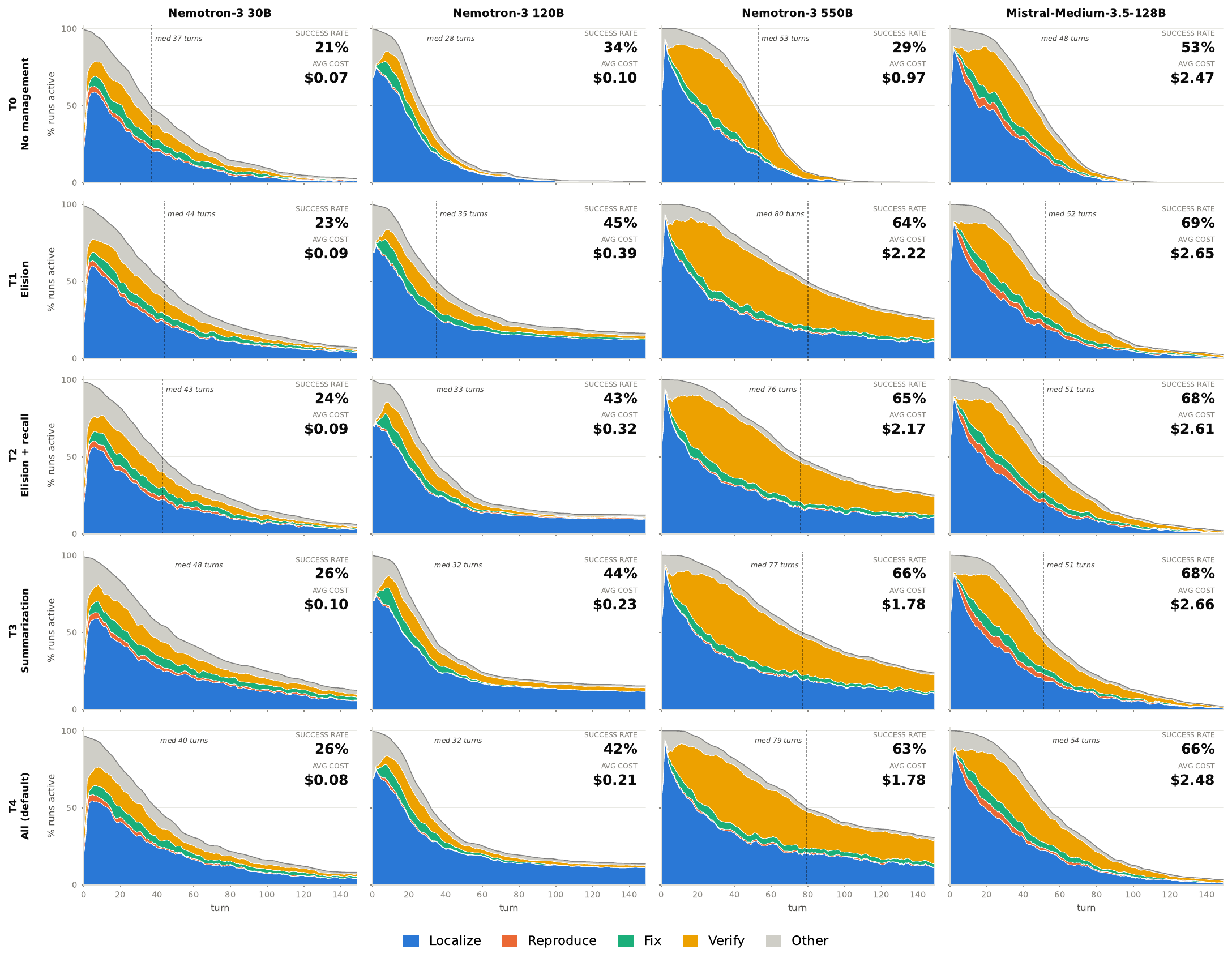}
\caption{SWE-Bench trajectory profiles across context-management strategies at 64k.}
\label{fig:swe_behavior_ctx64k}
\end{figure}

\begin{figure}[h]
\centering
\includegraphics[width=\textwidth]{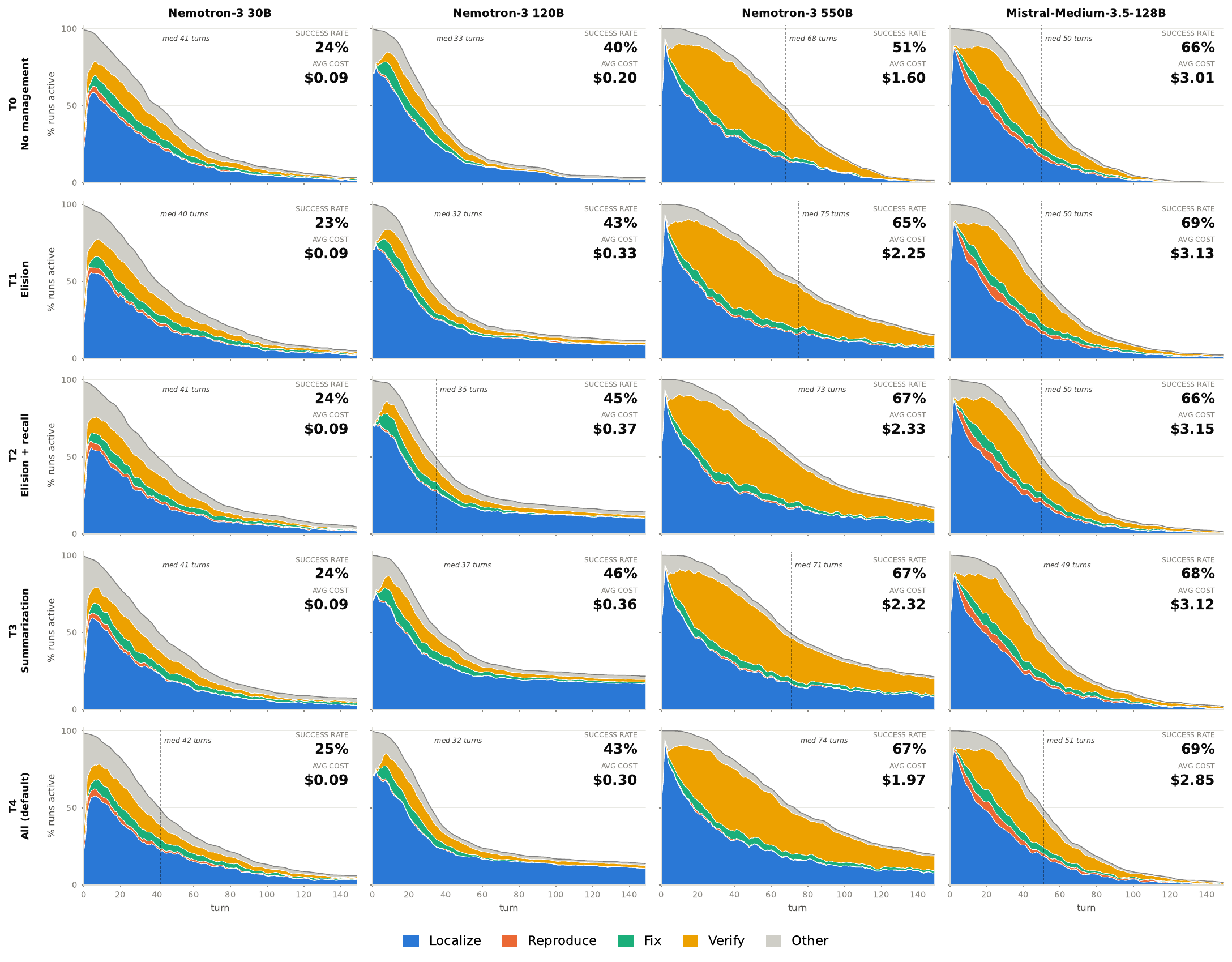}
\caption{SWE-Bench trajectory profiles across context-management strategies at 96k.}
\label{fig:swe_behavior_ctx96k}
\end{figure}

\begin{figure}[h]
\centering
\includegraphics[width=\textwidth]{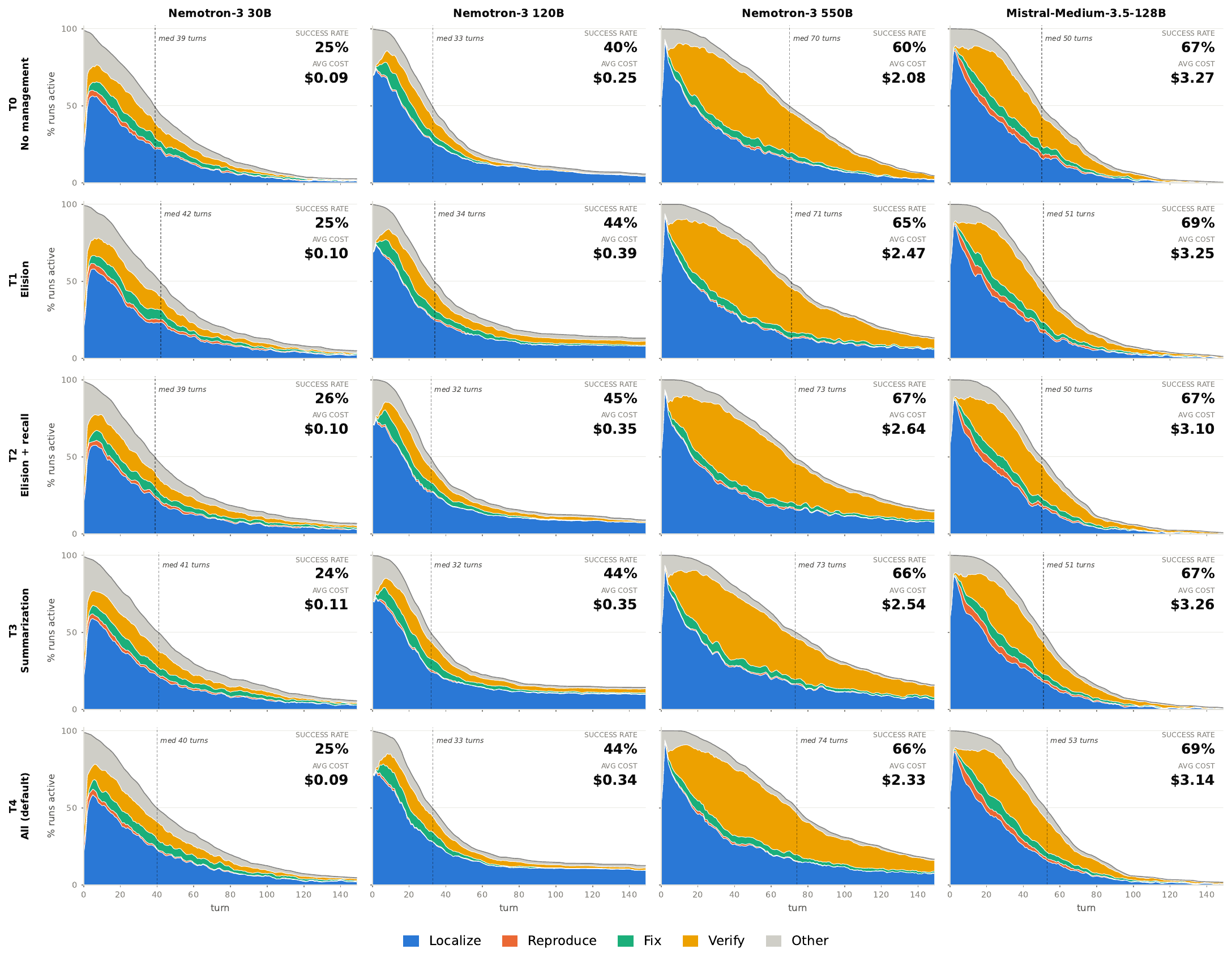}
\caption{SWE-Bench trajectory profiles across context-management strategies at 128k.}
\label{fig:swe_behavior_ctx128k}
\end{figure}

Figures~\ref{fig:swe_behavior_ctx32k}--\ref{fig:swe_behavior_ctx128k} hold planning and the predefined tool set fixed while varying the context-management tier and context-window budget. Columns correspond to models and rows correspond to T0--T4. At turn \(t\), the total stacked height is the percentage of trajectories that remain active, while the colored bands partition those active trajectories into Localize, Reproduce, Fix, Verify, and Other behavior. Dashed vertical lines indicate median trajectory length, and the upper-right annotations report success rate and mean cost per task. These profiles are descriptive summaries of the agent's behavior distribution.

\subsection{Terminal-Bench Trajectory Results}

\begin{figure}[h]
\centering
\includegraphics[width=\textwidth]{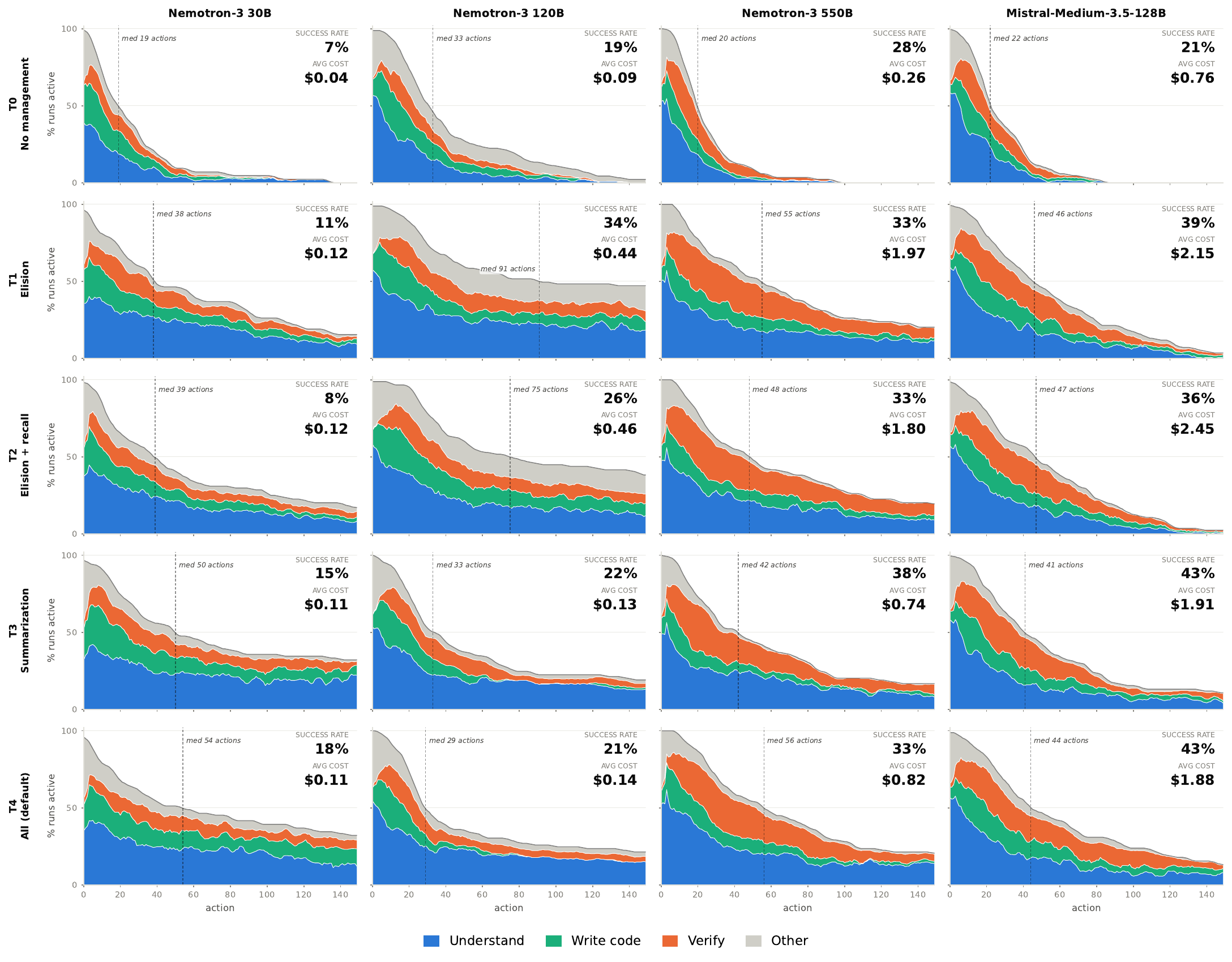}
\caption{Terminal-Bench trajectory profiles across context-management strategies at 32k.}
\label{fig:tb_behavior_ctx32k}
\end{figure}

\begin{figure}[h]
\centering
\includegraphics[width=\textwidth]{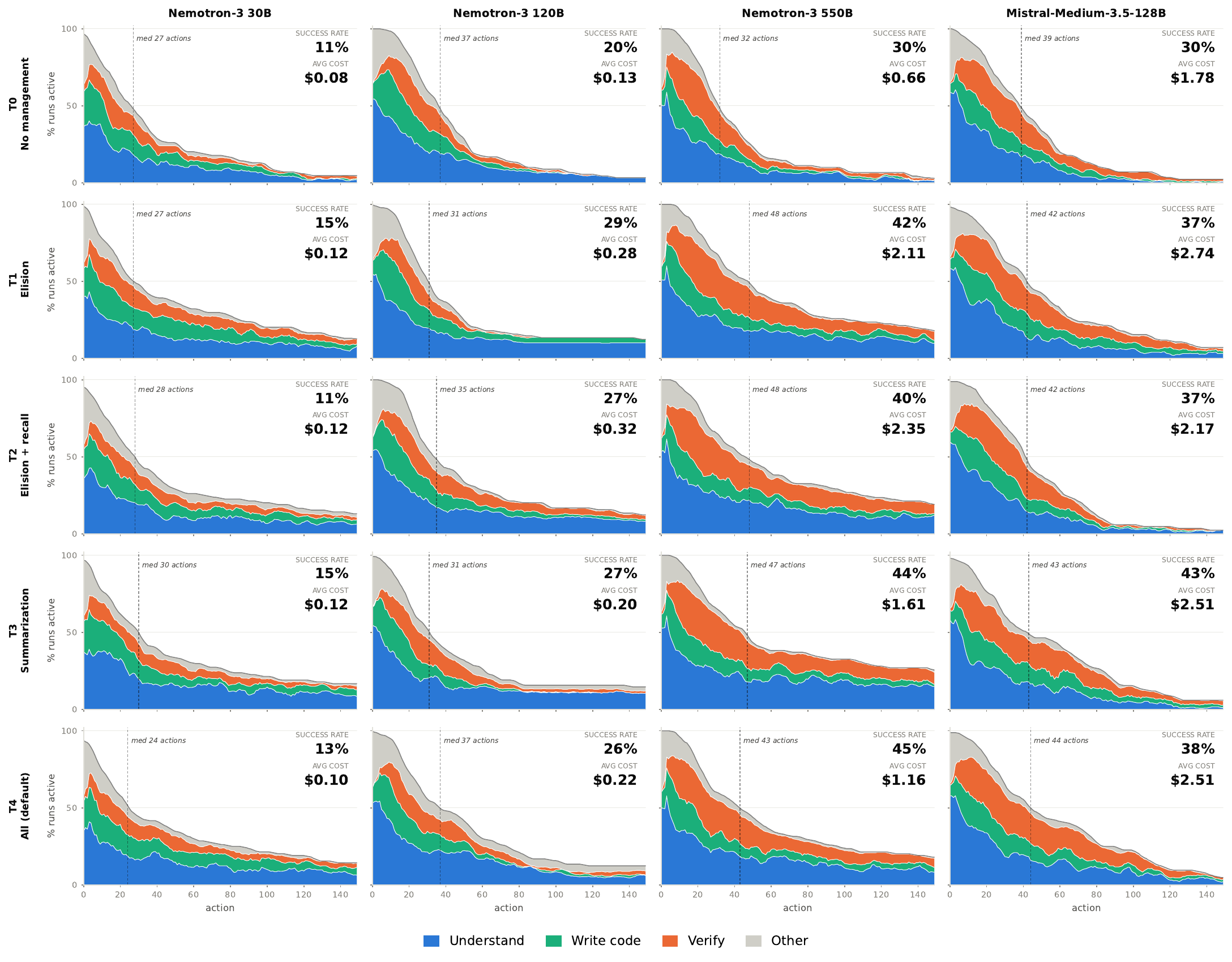}
\caption{Terminal-Bench trajectory profiles across context-management strategies at 64k.}
\label{fig:tb_behavior_ctx64k}
\end{figure}

\begin{figure}[h]
\centering
\includegraphics[width=\textwidth]{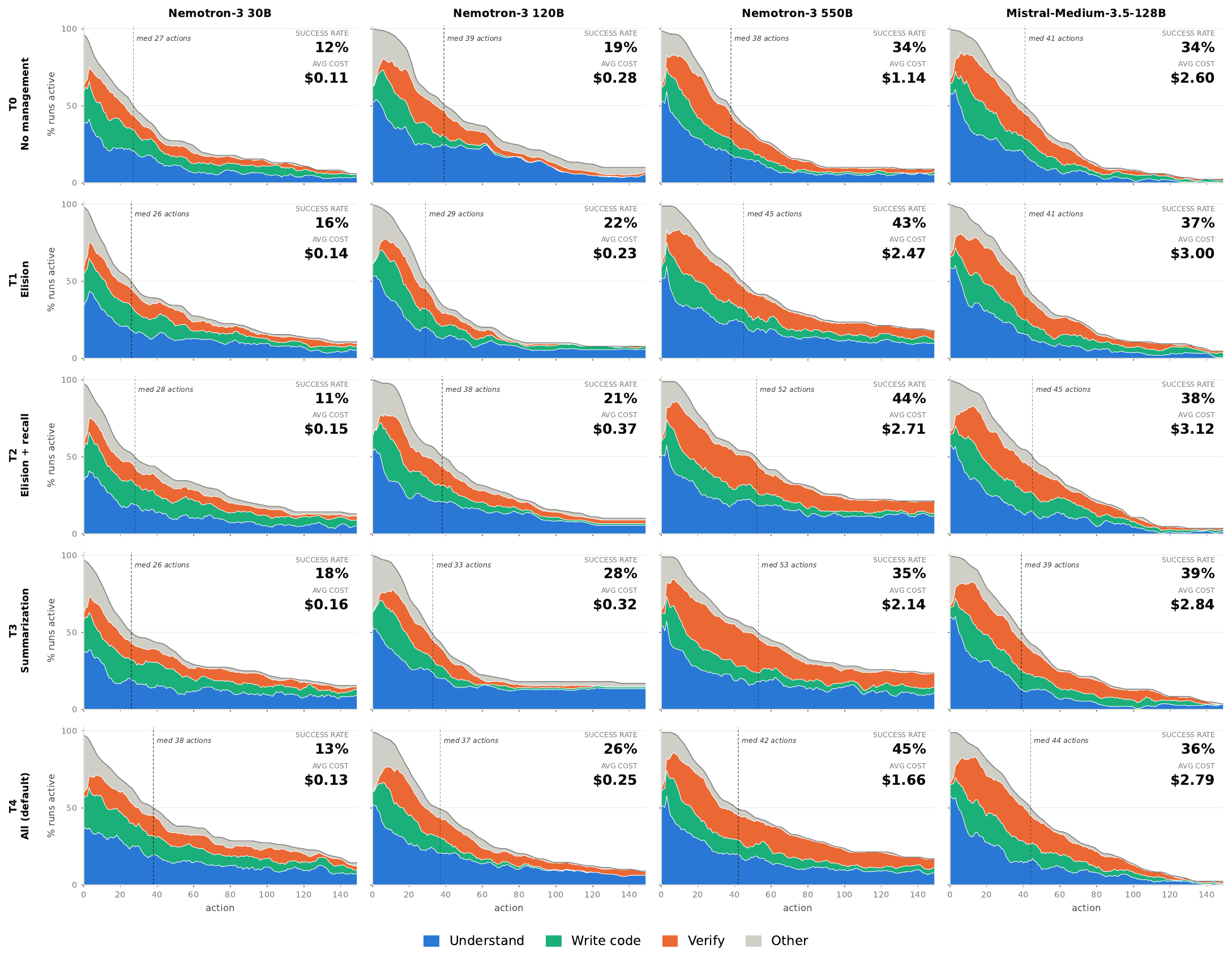}
\caption{Terminal-Bench trajectory profiles across context-management strategies at 96k.}
\label{fig:tb_behavior_ctx96k}
\end{figure}

\begin{figure}[h]
\centering
\includegraphics[width=\textwidth]{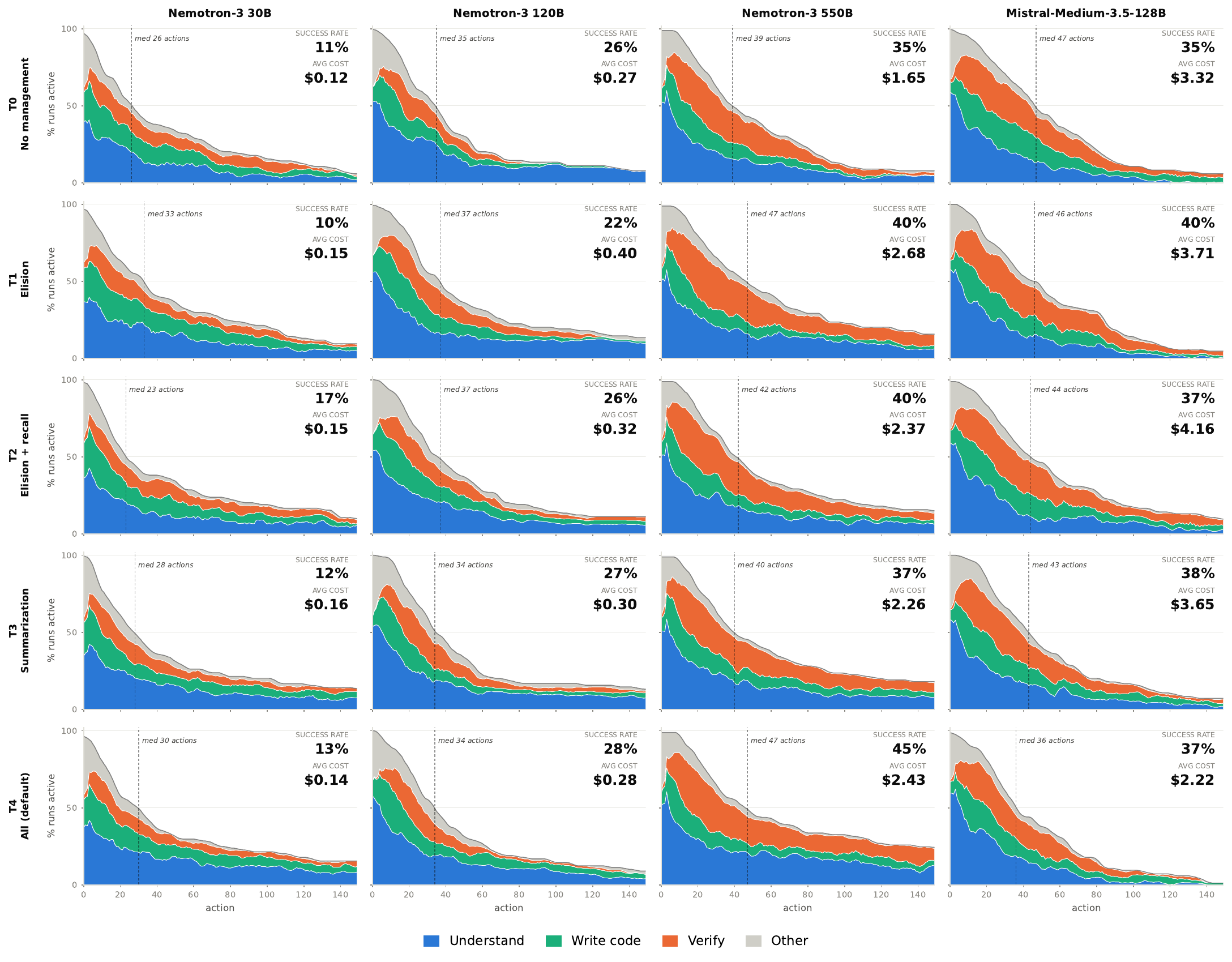}
\caption{Terminal-Bench trajectory profiles across context-management strategies at 128k.}
\label{fig:tb_behavior_ctx128k}
\end{figure}

Figures~\ref{fig:tb_behavior_ctx32k}--\ref{fig:tb_behavior_ctx128k} show the Terminal-Bench trajectory profiles, using the same layout as the SWE-Bench profiles. Columns correspond to models and rows correspond to T0--T4, with planning and the predefined tool set fixed. At turn \(t\), the total height reports the percentage of trajectories that remain active, while the colored bands partition active trajectories into Understand, Write Code, Verify, and Other behavior.

\subsection{Trajectory-Level Statistics at 128k}
Tables~\ref{tab:app_actionspace_granularity_full} and~\ref{tab:app_termination_stage_full} report the complete 128k statistics behind the trajectory-level summaries in the main text. Table~\ref{tab:app_actionspace_granularity_full} extends Table~\ref{tab:actionspace_granularity} with trajectory length, tool-call counts, and all five context-management tiers, while Table~\ref{tab:app_termination_stage_full} extends Table~\ref{tab:termination_stage} with the remaining tiers, the bash-only ablation, and the corresponding Terminal-Bench columns. The termination stages are read directly off the behavior encodings: on SWE-Bench, a run is counted as terminating without an edit if it contains no Fix turn, and as stalled at localization if, in addition, all of its labeled turns are Localize or Other turns; on Terminal-Bench, the analogous columns use the Write-code phase (Create or Modify actions) and the Understand phase (Inspect or Search actions) of the action-level encoding. 

\paragraph{Context-management tiers leave the trajectory shape almost unchanged at 128k.} Across T0--T4, the median trajectory length and the mean number of tool calls vary only within a narrow band for every model (Table~\ref{tab:app_actionspace_granularity_full}): for Nemotron-3 30B the SWE-Bench median stays between 39 and 42 turns, and for Nemotron-3 550B between 70 and 74 turns, while re-patch counts vary by less than two per task and median edit sizes by a few lines. Termination behavior is similarly stable (Table~\ref{tab:app_termination_stage_full}): the fraction of SWE-Bench runs that terminate without an edit varies across tiers by less than four points for three of the four models and by about eight points for Nemotron-3 120B. This stability is why \S\ref{sec:analysis} extends execution trajectories without substantially altering agent behavior at 128k; the tier-dependent differences appear only when the window binds, as in the 32k profiles (Figures~\ref{fig:swe_behavior_ctx32k} and~\ref{fig:tb_behavior_ctx32k}).

\paragraph{Planning and the action space are the interventions that change the trajectory.} Removing planning collapses the Nemotron-3 30B SWE-Bench trajectory to a median of 5 turns and pushes the without-edit termination rate from 27.8\% to 68.6\%, whereas for Nemotron-3 550B and Mistral-Medium-3.5-128B it lengthens the trajectory (from 74 to 108 and from 53 to 68 median turns) while leaving the without-edit rate below 3\%; these are the two regimes discussed in \S\ref{sec:analysis}. Bash-only reduces re-patching for all four models, but its effect on edit size is model-dependent: the median largest edit grows from 18 to 54 lines for Nemotron-3 550B, shrinks from 26 to 13 lines for Nemotron-3 30B, is essentially unchanged for Nemotron-3 120B, and decreases from 87 to 68 lines for Mistral-Medium-3.5-128B, whose edits are already large under the predefined tools. On Terminal-Bench, the create-or-replace share of file-writing actions rises under bash-only for every model, and the Nemotron-3 30B bash-only setting has the highest without-code termination rate of any configuration (22.5\%), consistent with the out-of-interface tool emissions described in \S\ref{sec:results}.

\begin{table}[t]
\centering
\caption{Full action-granularity statistics at a 128k context-window budget, corresponding to Table~\ref{tab:actionspace_granularity}. Rows include T0--T4 context-management tiers with planning and the full tool set, followed by T4 no-planning and bash-only ablations. \emph{Re-patch} counts edits to already-edited files, \emph{edit lines} is the median largest edit, and \emph{create share} is the fraction of Terminal-Bench file-writing actions that create or replace files.}
\label{tab:app_actionspace_granularity_full}
\footnotesize
\resizebox{\linewidth}{!}{%
\begin{tabular}{@{}llrrrrrrrr@{}}
\toprule
 & & \multicolumn{5}{c}{SWE-Bench Verified} & \multicolumn{3}{c}{Terminal-Bench} \\
\cmidrule(lr){3-7} \cmidrule(lr){8-10}
Model & Setting & SR (\%) & Med.\ turns & Calls & Re-patch & Edit lines & SR (\%) & Med.\ turns & Create (\%) \\
\midrule
\multirow{7}{*}{Nemotron-3 30B} & T0 & 24.8 & 39 & 48.0 & 2.0 & 22 & 11.2 & 26 & 26.6 \\
 & T1 & 25.0 & 42 & 54.5 & 2.7 & 28 & 10.1 & 34 & 23.0 \\
 & T2 & 26.0 & 39 & 54.4 & 2.9 & 27 & 16.9 & 23 & 27.8 \\
 & T3 & 23.6 & 41 & 56.0 & 3.3 & 24 & 12.4 & 28 & 34.3 \\
 & T4 & 25.2 & 40 & 54.6 & 3.3 & 26 & 13.5 & 30 & 27.5 \\
\cmidrule(lr){2-10}
 & T4 w/o plan & 13.6 & 5 & 9.8 & 0.7 & 7 & 9.0 & 20 & 24.0 \\
 & T4 bash only & 10.2 & 10 & 22.3 & 0.4 & 13 & 3.4 & 10 & 64.4 \\
\cmidrule(lr){1-10}
\multirow{7}{*}{Nemotron-3 120B} & T0 & 40.2 & 33 & 48.9 & 2.5 & 21 & 25.8 & 35 & 57.2 \\
 & T1 & 44.4 & 34 & 69.5 & 4.2 & 19 & 22.5 & 37 & 55.5 \\
 & T2 & 45.2 & 32 & 57.6 & 2.8 & 19 & 25.8 & 37 & 64.2 \\
 & T3 & 44.0 & 32 & 70.3 & 4.4 & 21 & 27.0 & 34 & 36.4 \\
 & T4 & 44.0 & 33 & 65.8 & 2.8 & 22 & 28.1 & 34 & 39.1 \\
\cmidrule(lr){2-10}
 & T4 w/o plan & 46.6 & 27 & 46.0 & 3.5 & 16 & 28.1 & 30 & 44.7 \\
 & T4 bash only & 42.4 & 41 & 84.6 & 2.2 & 24 & 23.6 & 40 & 75.9 \\
\cmidrule(lr){1-10}
\multirow{7}{*}{Nemotron-3 550B} & T0 & 59.8 & 70 & 77.0 & 3.1 & 18 & 34.8 & 39 & 53.8 \\
 & T1 & 65.2 & 71 & 91.7 & 3.9 & 19 & 40.4 & 47 & 31.5 \\
 & T2 & 67.4 & 73 & 96.4 & 4.4 & 19 & 40.4 & 42 & 58.2 \\
 & T3 & 65.8 & 74 & 95.8 & 4.2 & 18 & 37.1 & 40 & 58.6 \\
 & T4 & 65.8 & 74 & 98.4 & 4.6 & 18 & 44.9 & 47 & 50.9 \\
\cmidrule(lr){2-10}
 & T4 w/o plan & 67.8 & 108 & 130.3 & 4.6 & 19 & 46.1 & 38 & 45.3 \\
 & T4 bash only & 69.4 & 55 & 67.1 & 1.5 & 54 & 50.6 & 31 & 76.1 \\
\cmidrule(lr){1-10}
\multirow{7}{*}{Mistral-Medium-3.5-128B} & T0 & 67.4 & 50 & 56.6 & 2.8 & 86 & 34.8 & 47 & 34.7 \\
 & T1 & 68.6 & 51 & 56.6 & 2.8 & 86 & 40.5 & 46 & 37.1 \\
 & T2 & 67.0 & 50 & 55.6 & 2.7 & 87 & 37.1 & 44 & 31.7 \\
 & T3 & 66.6 & 51 & 57.1 & 2.8 & 88 & 38.2 & 44 & 22.2 \\
 & T4 & 68.6 & 53 & 57.8 & 3.0 & 87 & 37.1 & 36 & 27.9 \\
\cmidrule(lr){2-10}
 & T4 w/o plan & 69.0 & 68 & 75.3 & 3.3 & 94 & 39.3 & 48 & 37.3 \\
 & T4 bash only & 45.4 & 47 & 50.2 & 1.3 & 68 & 43.8 & 43 & 57.0 \\
\bottomrule
\end{tabular}%
}
\end{table}

\begin{table}[t]
\centering
\caption{Full termination-stage statistics at a 128k context-window budget, corresponding to Table~\ref{tab:termination_stage}. The termination columns are derived from the behavior encodings in Tables~\ref{tab:traj_encoding} and~\ref{tab:tbench_purpose}: \emph{W/o edit} (\emph{W/o code}) is the fraction of runs that terminate without any Fix turn (without any Create or Modify action), \emph{At loc.}\ (\emph{At underst.}) is the subset of those runs that never progress beyond the Localize phase (the Understand phase, i.e., Inspect and Search actions), apart from Other-phase bookkeeping, and \emph{Edited} (\emph{Wrote}) is the fraction of runs that did edit source (write code) but remained unresolved.}
\label{tab:app_termination_stage_full}
\small
\resizebox{\linewidth}{!}{%
\begin{tabular}{@{}llrrrrrrrr@{}}
\toprule
 & & \multicolumn{4}{c}{SWE-Bench Verified} & \multicolumn{4}{c}{Terminal-Bench} \\
\cmidrule(lr){3-6} \cmidrule(lr){7-10}
 & & \multicolumn{2}{c}{Terminated (\%)} & & & \multicolumn{2}{c}{Terminated (\%)} & & \\
\cmidrule(lr){3-4} \cmidrule(lr){7-8}
Model & Setting & W/o edit & At loc. & Edited & SR (\%)  & W/o code & At underst. & Wrote & SR (\%)  \\
\midrule
\multirow{7}{*}{Nemotron-3 30B} & T0 & 26.4 & 9.6 & 48.8 & 24.8 & 7.1 & 7.1 & 81.0 & 11.2 \\
 & T1 & 24.6 & 7.0 & 51.0 & 25.0 & 11.9 & 7.1 & 77.4 & 10.1 \\
 & T2 & 27.6 & 10.4 & 46.0 & 26.0 & 3.6 & 3.6 & 79.8 & 16.9 \\
 & T3 & 26.2 & 8.2 & 50.2 & 23.6 & 8.3 & 3.6 & 78.6 & 12.4 \\
 & T4 & 27.8 & 10.4 & 47.0 & 25.2 & 8.3 & 6.0 & 77.4 & 13.5 \\
\cmidrule(lr){2-10}
 & T4 w/o plan & 68.6 & 58.4 & 18.0 & 13.6 & 10.7 & 4.8 & 81.0 & 9.0 \\
 & T4 bash only & 75.4 & 47.6 & 15.0 & 10.2 & 22.5 & 14.6 & 74.2 & 3.4 \\
\cmidrule(lr){1-10}
\multirow{7}{*}{Nemotron-3 120B} & T0 & 18.6 & 16.2 & 41.4 & 40.2 & 11.2 & 6.7 & 62.9 & 25.8 \\
 & T1 & 16.6 & 11.8 & 39.6 & 44.4 & 11.2 & 6.7 & 66.3 & 22.5 \\
 & T2 & 21.6 & 13.0 & 33.4 & 45.2 & 11.2 & 5.6 & 62.9 & 25.8 \\
 & T3 & 20.2 & 13.4 & 36.2 & 44.0 & 12.4 & 4.5 & 60.7 & 27.0 \\
 & T4 & 24.4 & 17.8 & 31.8 & 44.0 & 13.5 & 9.0 & 58.4 & 28.1 \\
\cmidrule(lr){2-10}
 & T4 w/o plan & 15.6 & 11.0 & 38.4 & 46.6 & 10.1 & 6.7 & 61.8 & 28.1 \\
 & T4 bash only & 24.4 & 12.4 & 33.4 & 42.4 & 10.1 & 6.7 & 66.3 & 23.6 \\
\cmidrule(lr){1-10}
\multirow{7}{*}{Nemotron-3 550B} & T0 & 1.8 & 1.6 & 38.4 & 59.8 & 7.9 & 2.2 & 57.3 & 34.8 \\
 & T1 & 2.2 & 0.6 & 32.6 & 65.2 & 3.4 & 1.1 & 56.2 & 40.4 \\
 & T2 & 2.4 & 0.4 & 30.2 & 67.4 & 5.6 & 2.2 & 53.9 & 40.4 \\
 & T3 & 2.0 & 0.4 & 32.2 & 65.8 & 2.2 & 1.1 & 60.7 & 37.1 \\
 & T4 & 2.4 & 0.6 & 31.8 & 65.8 & 2.2 & 1.1 & 52.8 & 44.9 \\
\cmidrule(lr){2-10}
 & T4 w/o plan & 1.4 & 0.2 & 30.8 & 67.8 & 4.5 & 0.0 & 49.4 & 46.1 \\
 & T4 bash only & 1.6 & 0.4 & 29.0 & 69.4 & 3.4 & 1.1 & 46.1 & 50.6 \\
\cmidrule(lr){1-10}
\multirow{7}{*}{Mistral-Medium-3.5-128B} & T0 & 1.6 & 0.4 & 31.4 & 67.4 & 8.3 & 2.4 & 54.8 & 34.8 \\
 & T1 & 1.8 & 0.6 & 29.6 & 68.6 & 4.8 & 1.2 & 53.6 & 40.5 \\
 & T2 & 2.2 & 0.8 & 31.0 & 67.0 & 6.0 & 2.4 & 57.1 & 37.1 \\
 & T3 & 2.4 & 1.0 & 31.2 & 66.6 & 6.0 & 0.0 & 53.6 & 38.2 \\
 & T4 & 1.2 & 0.6 & 30.2 & 68.6 & 15.5 & 4.8 & 46.4 & 37.1 \\
\cmidrule(lr){2-10}
 & T4 w/o plan & 2.0 & 1.0 & 29.2 & 69.0 & 8.3 & 6.0 & 51.2 & 39.3 \\
 & T4 bash only & 32.8 & 17.8 & 22.0 & 45.4 & 7.9 & 3.4 & 48.3 & 43.8 \\
\bottomrule
\end{tabular}%
}
\end{table}

\subsection{Recall-Event Usage}
\label{app:recall_usage}

Table~\ref{tab:recall_usage} reports the complete per-configuration invocation rates used in \S\ref{sec:recall_usage}. The reported statistic is the mean number of \texttt{recall\_event} invocations per task rather than the fraction of trajectories that use recall, because a trajectory may invoke the tool multiple times.

\begin{table}[t]
    \centering
    \small
    \setlength{\tabcolsep}{4pt}
    \caption{Mean number of \texttt{recall\_event} calls per task in the recall-enabled context-management strategies (T2 and T4), by model and context-window budget. All 16 planning and action-space ablation configurations at T4/128k record zero calls and are omitted.}
    \label{tab:recall_usage}
    \begin{tabular*}{\textwidth}{@{\extracolsep{\fill}}llrrrrrrrr@{}}
        \toprule
        & & \multicolumn{4}{c}{SWE-Bench Verified}
        & \multicolumn{4}{c}{Terminal-Bench} \\
        \cmidrule(r{6pt}){3-6}\cmidrule(l){7-10}
        Model & Tier
        & 32k & 64k & 96k & 128k
        & 32k & 64k & 96k & 128k \\
        \midrule
        Nemotron-3 30B
            & T2 & 0.146 & 0.004 & 0.000 & 0.000 & 4.326 & 0.674 & 0.000 & 0.045 \\
            & T4 & 0.472 & 0.032 & 0.002 & 0.002 & 2.708 & 0.360 & 0.169 & 0.067 \\
        \addlinespace
        Nemotron-3 120B
            & T2 & 0.206 & 0.008 & 0.004 & 0.000 & 0.250 & 0.000 & 0.000 & 0.000 \\
            & T4 & 0.318 & 0.010 & 0.002 & 0.000 & 0.022 & 0.000 & 0.000 & 0.000 \\
        \addlinespace
        Nemotron-3 550B
            & T2 & 0.000 & 0.000 & 0.000 & 0.000 & 0.022 & 0.000 & 0.000 & 0.000 \\
            & T4 & 0.006 & 0.000 & 0.000 & 0.000 & 0.090 & 0.000 & 0.000 & 0.000 \\
        \addlinespace
        Mistral-Medium-3.5-128B
            & T2 & 0.000 & 0.000 & 0.000 & 0.000 & 0.022 & 0.000 & 0.000 & 0.000 \\
            & T4 & 0.010 & 0.000 & 0.000 & 0.000 & 0.045 & 0.011 & 0.000 & 0.000 \\
        \bottomrule
    \end{tabular*}
\end{table}

\subsection{Judge Prompts}
Figures~\ref{prompt:turn_purposes}, \ref{prompt:failure_stage}, and~\ref{prompt:action_purpose} reproduce the full prompts used for SWE-Bench turn-purpose classification, SWE-Bench failure-stage attribution, and Terminal-Bench action-purpose classification.
\begin{figure}[p]
\centering
\begin{tcolorbox}[
    colback=background,
    colframe=frame,
    width=\textwidth,
    left=4pt, right=4pt, top=2pt, bottom=2pt, boxsep=1pt
]
\begin{Verbatim}[breaklines=true,breaksymbol=,fontsize=\scriptsize]
<instruction>
You are analyzing a software debugging trajectory where an agent attempts to fix a bug.
The trajectory is divided into turns (model invocations), where each turn may execute
one or more tool calls (reading files, editing code, running tests, etc.).

For EACH turn, classify its PRIMARY workflow purpose into exactly one category:

## Purpose categories

- `L` = **Localize**: Finding the bug location through reading, searching,
                      or grepping files. The goal is to understand WHERE
                      the bug is in the codebase.

- `R` = **Reproduce**: Observing or reproducing the buggy behavior by running
                       scripts, executing code, or checking outputs. The goal
                       is to confirm WHAT the bug does.

- `F` = **Fix**: Editing source files to implement a fix. This includes
                 any modifications to actual project source code (not
                 scratch/repro files).
                 IMPORTANT: Only classify as F if the edit RESOLVES the bug.
                 Files created/written to reproduce, test, or verify the bug
                 are NOT fixes (use R or V instead).

- `V` = **Verify**: Running tests or validation to check if the fix works.
                    This includes pytest, unittest, or any test execution.
                    Also includes CREATING new test files to verify the bug or fix.

- `O` = **Other**: Everything else — environment setup (pip/apt), git
                   operations, file management, navigation (cd/pwd),
                   or miscellaneous tasks.

## Classification rules

- If a turn has multiple actions with DIFFERENT purposes, pick the MOST IMPORTANT one:
  * Fix (F) is highest priority — if the turn edits source, it's an F turn
  * Verify (V) comes next — if testing happens, it's likely a V turn
  * Reproduce (R) and Localize (L) are exploration activities
  * Other (O) is the fallback

- Judge by the SEMANTIC INTENT, not just the tool names:
  * Reading a file to understand code structure = L (Localize)
  * Reading test output to see if tests pass = V (Verify)
  * Running a reproduction script = R (Reproduce)
  * Creating repro.py / test_bug.py to reproduce the issue = R (Reproduce)
  * Creating test_fix.py to verify the fix = V (Verify)
  * Editing source to change behavior and resolve the bug = F (Fix)
</instruction>

<trajectory>
{turns_json}
</trajectory>

<output_format>
There are exactly {n_turns} turns above, numbered "Turn 1" through "Turn {n_turns}".

Respond with ONLY a JSON object mapping EVERY turn number (as a string key)
to its single-letter purpose. You MUST include a key for every turn from
"1" to "{n_turns}" — do not skip any turn.

Example format (for 5 turns):
{"1": "L", "2": "R", "3": "F", "4": "V", "5": "O"}

No preamble, no commentary, no code fence — JUST the JSON object.
</output_format>
\end{Verbatim}
\end{tcolorbox}
\caption{LLM-judge prompt used for \textbf{SWE-Bench trajectory encoding}. The judge assigns each turn to one of the five phase symbols in Table~\ref{tab:traj_encoding}.}
\label{prompt:turn_purposes}
\end{figure}

\begin{figure}[p]
\centering
\begin{tcolorbox}[
    colback=background,
    colframe=frame,
    width=\textwidth,
    left=4pt, right=4pt, top=2pt, bottom=2pt, boxsep=1pt
]
\begin{Verbatim}[breaklines=true,breaksymbol=,fontsize=\scriptsize]
<instruction>
You are diagnosing WHERE in the problem-solving pipeline a coding agent broke
down on a software issue it FAILED to resolve. The pipeline runs in this order:

  file_localization → line_localization → patch_implementation → verification

Pick the SINGLE EARLIEST stage the agent did NOT complete correctly — the stage
where the trajectory first went off the rails. Everything before that stage was
done adequately; the failure is rooted here.

## Stage definitions

- `file_localization`: never opened / edited the file(s) that actually need to
                       change (compare against the gold-changed files).
- `line_localization`: found the right file(s) but never focused on the right
                       region/function/lines within them.
- `patch_implementation`: wrote an incorrect, incomplete, or overly narrow fix.
- `verification`: the fix is essentially on-target but the agent failed at
                  testing/validation — never ran the tests, misread their
                  result, or shipped a change that breaks other tests.

## Allowed labels

Pick exactly one label from this set:
{labels}
</instruction>

<issue>
{issue}
</issue>

<gold_changed_files>
{gold_files}
</gold_changed_files>

<agent_patch>
{patch}
</agent_patch>

<agent_transcript>
{transcript}
</agent_transcript>

<output_format>
Respond with ONLY a JSON object — no preamble, commentary, or code fence:
{"label": "<one stage>", "reason": "<short: what was done up to here and what broke>"}
</output_format>
\end{Verbatim}
\end{tcolorbox}
\caption{LLM-judge prompt used for \textbf{SWE-Bench failure-stage attribution}. For each unresolved trajectory, the judge assigns the earliest failed stage in the repair pipeline: file localization, line localization, patch implementation, or verification. }
\label{prompt:failure_stage}
\end{figure}

\begin{figure}[p]
\centering
\begin{tcolorbox}[
    colback=background,
    colframe=frame,
    width=\textwidth,
    left=4pt, right=4pt, top=2pt, bottom=2pt, boxsep=1pt
]
\begin{Verbatim}[breaklines=true,breaksymbol=,fontsize=\scriptsize]
<instruction>
You are analyzing the execution trace of a coding agent working on a Terminal Bench task. The agent works inside a Linux container: it reads and writes files, installs packages, and runs shell commands until it believes the task is done.

The trajectory below is a sequence of TURNS (model invocations); each turn executes one or more TOOL CALLS (actions).

Classify the PURPOSE of EACH individual action (each tool call) into exactly one of the following codes. Judge by the SEMANTIC INTENT of the action, not merely the tool name or the leading shell token.

## Purpose codes
- `I`  Inspect      Read file CONTENT to understand it (read_file, cat, head, tail, hexdump, od, strings).
- `S`  Search       Discover WHAT EXISTS: list a directory or search for files/text (list_files, ls, find, grep, glob).
- `C`  Create       Author a NEW file — script, config, or output artifact (write_file to a fresh path).
- `M`  Modify       Change an EXISTING file in place (edit_file, sed -i, patch, rewriting a file the agent already wrote).
- `E`  Environment  Install or build dependencies (pip/apt/conda/npm install, make, cmake, ./configure).
- `X`  eXecute      Run the agent's OWN work to produce a result
- `T`  Test         Invoke a real test framework (pytest, unittest, cargo test, npm test, go test).
- `V`  Verify       Check an ALREADY-PRODUCED artifact against the task's stated success condition.
- `N`  Navigate     Move around the filesystem, where that is the ENTIRE point of the call (bare cd, pwd).
- `G`  General      Anything else: git, echo, shell plumbing (awk/sed/sort over intermediate data), cleanup, misc. update plan must be considered as General (G).

## Rules
- Judge intent, not the leading token. Classify what the action is FOR, not which binary appears first. `cd /app && python3 solve.py` is X (running the solution) — the `cd` is incidental scaffolding, not the purpose. Only use N when moving around the filesystem is the ENTIRE point of the call, e.g. a bare `cd /app` or `pwd` with nothing chained to it.
- Classify EACH action independently; a single turn may contain actions with different purposes. Do not smear one label across a turn.
- X vs T vs V. X runs the agent's own work to PRODUCE a result — its analysis script, its solution, an inline heredoc computation. T invokes a genuine test framework (pytest, unittest, cargo test, npm test, go test), including the task's provided test file. V CHECKS an artifact that already exists against the task's stated success condition — re-reading the written answer file to confirm its format, diffing produced output against an expected value. When a single run both produces and checks a result, prefer X.
- C vs M. C authors a NEW file at a path not written before in this trajectory. M changes something that already exists — `edit_file`, `sed -i`, `patch`, or `write_file` overwriting a path the agent itself wrote earlier. A rewritten-from-scratch v2 of the agent's own script at a NEW filename (analyze2.py after analyze.py) is still C.
- C vs X for heredocs. `python3 << 'EOF' ... EOF` and `bash -c '...'` EXECUTE code; they do not author a file. Those are X. Only a real write to a path is C.
- E is setup, not building the answer. E covers making the environment usable: pip/apt/conda/npm install, make, cmake, ./configure, compiling a third-party dependency. Compiling the agent's OWN program as the deliverable is X.
- Errors do not change the purpose. An action marked ERR is labelled by what it was TRYING to do. A failed `pip install` is still E; a crashed solution script is still X.
</instruction>

<trajectory>
{actions_block}
</trajectory>

<output_format>
There are exactly {n_actions} actions above, numbered "[1]" through "[{n_actions}]".

Respond with ONLY a JSON object holding one key per ACTION, named "Turn1" through "Turn{n_actions}". The number in the key is the ACTION number in square brackets above — NOT the "-- Turn N --" header, which only groups them. Each value is an object with exactly two fields, IN THIS ORDER:
  "Reason"  one short sentence (at most 20 words) saying what THIS action is for. Cite the concrete command, path or tool that decides it; do not just restate a code's definition. Where a rule above applies (intent-over-token, X vs T vs V, C vs M, heredoc, E-is-setup), name the rule that settles it.
  "Label"   the purpose code the reason leads to: one of "I", "S", "C", "C", "M", "E", "X", "T", "V", "N", "G".

Write "Reason" FIRST and let it decide "Label" — do not pick a code and then justify it.

You MUST include a key for every action from "Turn1" to "Turn{n_actions}" — do not skip any, and do not add keys beyond "Turn{n_actions}".

Example format (for 3 actions):
{"Turn1": {"Reason": "ls /app lists the directory to discover what files exist", "Label": "S"}, "Turn2": {"Reason": "head -20 data.csv reads real content to understand it, not a yes/no probe", "Label": "I"}, "Turn3": {"Reason": "runs the agent's own solve.py to produce the answer; the cd is incidental", "Label": "X"}}

No preamble, no commentary, no code fence — JUST the JSON object.
</output_format>
\end{Verbatim}
\end{tcolorbox}
\caption{LLM-judge prompt used for Terminal-Bench action-level encoding. The judge assigns each shell-based action to one of the 10 fine-grained action-type symbols in Table~\ref{tab:tbench_purpose}.}
\label{prompt:action_purpose}
\end{figure}

\end{document}